\pdfoutput=1

\documentclass{article} % For LaTeX2e

\usepackage{iclr2023_conference,times}
\usepackage[hidelinks]{hyperref}
\usepackage{url}
\usepackage{amsmath}
\usepackage[capitalize]{cleveref}
\usepackage{graphicx}
\usepackage{subcaption}
\usepackage{amssymb}
\usepackage{xspace}
\usepackage{multirow}
\usepackage{booktabs}
\usepackage{natbib}
\usepackage{mathabx}
\usepackage{scalerel}
\usepackage{graphicx}
\newcommand{\nback}[1][-.95pt]{
  \mathrel{\raisebox{#1}{$\rotatebox[origin=c]{-315}{\scaleobj{0.55}{-}}$}}
}
\newcommand{\undernegpreccurlyeq}{%
\mathrel{\ooalign{$\preccurlyeq$\cr\kern1.2pt$\nback$}}}
\usepackage{algorithm}
\usepackage{algorithmic}
\usepackage[english]{babel}
\usepackage{amsthm}
\theoremstyle{definition}
\newtheorem{definition}{Definition}[section]
\newcommand{\R}{\mathbb{R}}

\newcommand{\cN}{\mathcal{N}\xspace}
\newcommand{\cI}{\mathcal{I}\xspace}
\newcommand{\cC}{\mathcal{O}\xspace}

\newcommand{\cP}{\mathcal{P}\xspace}

\newcommand{\bW}{\boldsymbol{W}}
\newcommand{\bb}{\boldsymbol{b}}
\newcommand{\pertx}{B(x, \epsilon)}
\newcommand{\NAP}{\text{NAP}\xspace}
\newcommand{\Y}{\color{blue}Y}

\let\cite=\citep

\title{Towards reliable neural specifications}

\usepackage{authblk}

\author[1*]{Chuqin Geng}
\author[2*]{Nham Le}
\author[1]{Xiaojie Xu}
\author[1]{Zhaoyue Wang}
\author[2]{Arie Gurfinkel}
\author[1]{Xujie Si}
\affil[1]{Department of Computer Science, McGill University}
\affil[2]{Department of Electrical and Computer Engineering, University of Waterloo}
\affil[ ]{\texttt {\{chuqin.geng,xiaojie.xu,zhaoyue.wang\}@mail.mcgill.ca, xsi@cs.mcgill.ca, \{nv3le, arie.gurfinkel\}@uwaterloo.ca}}
\footnotetext{These authors contributed equally to this work.}
\iclrfinalcopy
\begin{document}

\maketitle

\begin{abstract}

Having reliable specifications is an unavoidable challenge in achieving verifiable correctness, robustness, and interpretability of AI systems. 
Existing specifications for neural networks are in the paradigm of \textit{data as specification}. That is, the local neighborhood centering around a reference input is considered to be correct (or robust). While existing specifications contribute to verifying adversarial robustness, a significant problem in many research domains, our empirical study shows that those verified regions are somewhat tight, and thus fail to allow verification of test set inputs, making them impractical for some real-world applications. 
To this end, we propose a new family of specifications called \textit{neural representation as specification}, which uses the intrinsic information of neural networks  --- neural activation patterns (NAPs), rather than input data to specify the correctness and/or robustness of neural network predictions.
%\NL{Should it be ``together with input data''? If we have experiments with perturbations surrounding an input, then we are using both the input specs and the NAP specs, not just the NAP.} 
We present a simple statistical approach to mining neural activation patterns. To show the effectiveness of discovered NAPs, we formally verify several important properties, such as various types of misclassifications will never happen for a given NAP, and there is no ambiguity between different NAPs. We show that by using NAP, we can verify a significant region of the input space, while still recalling 84\% of the data on MNIST. Moreover, we can push the verifiable bound to 10 times larger on the CIFAR10 benchmark. Thus, we argue that NAPs can potentially be used as a more reliable and extensible specification for neural network verification.

%\NL{Do we need to explain ``dominant''?}
% We analyze NAPs from a statistical point of view and find that a single NAP can cover a large number of training and testing data points whereas ad hoc data-as-specification only covers the given reference data point.

% To show the effectiveness of discovered NAPs, we formally verify several important properties, such as  various types of misclassifications will never happen for a given NAP, and there is no-ambiguity between different NAPs. We show that by using NAP, we can verify a significant region ofthe input space, while still recalling 84\% of the data on the MNIST dataset.
% Thus, we argue that NAPs could potentially be used as a more reliable and extensible specification for neural network verification.

% \xx{We analyze NAPs from a statistical point of view and find that a single NAP can cover a large number of training and testing data points which, proved by our experiments, is more effective than the data-as-specification as it only covers the given reference data point. }

% \begin{itemize}
%     \item activation patterns is a more useful/general/reliable way to give specifications
%     \item a simple approach to find dominant activation patterns
%     \item verifiable properties of activation patterns (certain type of misclassification never happen, no-ambiguity, etc.
% \end{itemize}

\end{abstract}

\section{Introduction}

The advances in deep neural networks (DNNs) have brought a wide societal impact in many domains such as transportation, healthcare, finance, e-commerce, and education. This growing societal-scale impact has also raised some risks and concerns about errors in AI software, their susceptibility to cyber-attacks, and AI system safety~\cite{DBLP:journals/cacm/DietterichH15}.
Therefore, the challenge of verification and validation of AI systems, as well as, achieving trustworthy AI~\cite{DBLP:journals/cacm/Wing21}, has attracted much attention of the research community. Existing works approach this challenge by building on \emph{formal methods} -- a field of computer science and engineering that involves verifying properties of systems using rigorous mathematical specifications and proofs~\cite{DBLP:journals/computer/Wing90}. 
Having a formal specification --- a precise, mathematical statement of what AI system is supposed to do is critical for formal verification. Most works~\cite{reluplex, Marabou, huang2017cav, huang2020csr, abc} use the specification of adversarial robustness for classification tasks that states that the NN correctly classifies an image as a given adversarial label under perturbations with a specific norm (usually $L_\infty$). Generally speaking, existing works use a paradigm of \textit{data as specification} --- the robustness of local neighborhoods of reference data points with ground-truth labels 
% (e.g., the training dataset) 
is the only specification of correct behaviors. 
% Similar to memorization 
However, from a learning perspective, this would lead to \textit{overfitted} specification, since only local neighborhoods of reference inputs get certified.

\begin{figure}[t]
     \centering
     \begin{subfigure}[t]{0.24\textwidth}
         \centering
         \includegraphics[width=\textwidth]{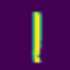}
         \caption{A testing image from MNIST, classified as 1}
         \label{fig:unseen}
     \end{subfigure}
     \hfill
     \begin{subfigure}[t]{0.24\textwidth}
         \centering
         \includegraphics[width=\textwidth]{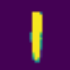}
         \caption{The closest training image in MNIST, whose $L_\infty$ distance 
         to Fig.~\ref{fig:unseen} is 0.5294}
         \label{fig:anchor}
     \end{subfigure}
     \hfill
     \begin{subfigure}[t]{0.24\textwidth}
         \centering
         \includegraphics[width=\textwidth]{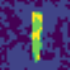}
         \caption{An adversarial example misclassified as 8, whose $L_\infty$ distance to Fig.~\ref{fig:anchor} is 0.2}
         \label{fig:adv_example}
     \end{subfigure}
     \hfill
      \begin{subfigure}[t]{0.24\textwidth}
         \centering
         \includegraphics[width=\textwidth]{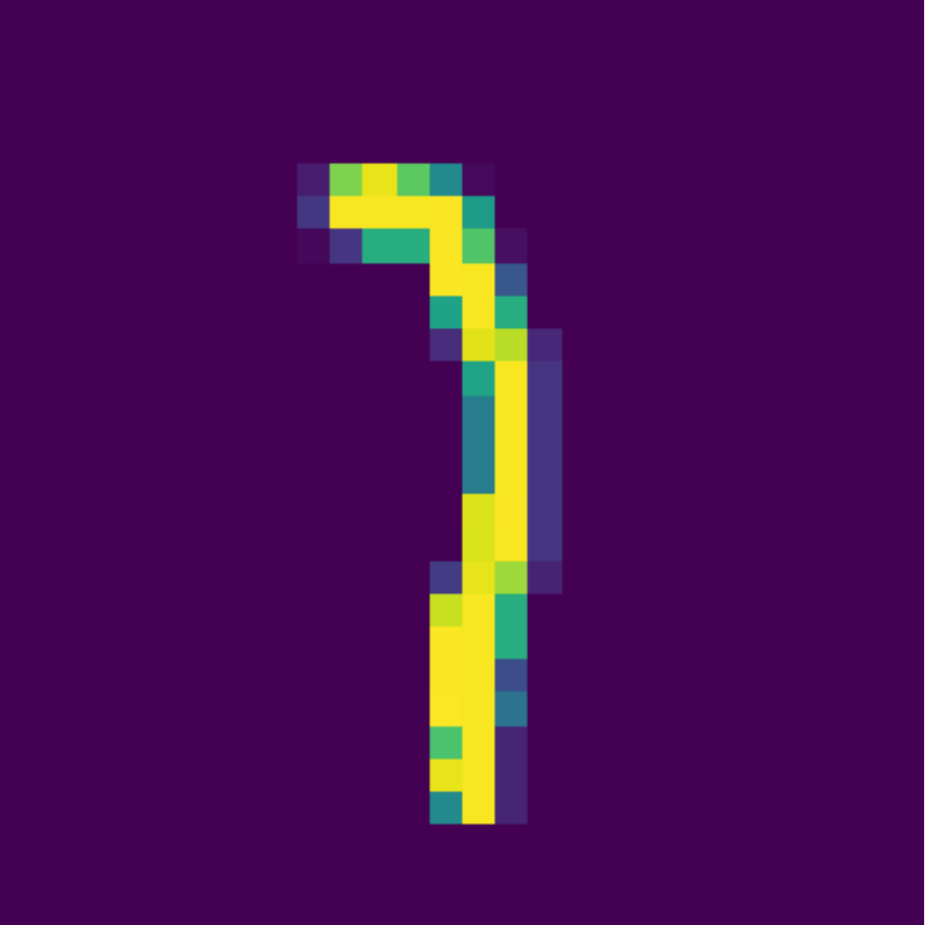}
         \caption{A testing image, on which our verified NAP (for digit 1) disagrees with the ground truth (i.e., 7)}
         \label{fig:nap_disagree}
         
      \end{subfigure}
        \caption{The limitation of ``data-as-specification'': First three images show that a test input can be much further away (in $L_\infty$) from its closest train input compared to adversarial examples (the upper bound of a verifiable local region). The last image shows that even data itself can be imperfect.}
        \label{fig:failure of distance}
\end{figure}

%This gap is problematic. \textbf{First}, only small regions ($L_\infty$ norm-balls) centered at reference points can be indeed verified using state-of-the-art solvers. Since both training and testing data are sampled from an underlying distribution, the chance of test data falling into those verifiable regions over the input space is quite small, making such verification less applicable to real-world problems. 

As a concrete example, \Cref{fig:failure of distance} illustrates the fundamental limitation of such overfitted specifications. Specifically, a testing input like the one shown in \cref{fig:unseen} can hardly be verified even if all local neighborhoods of all training images have been certified using the $L_\infty$ norm.
This is because adversarial examples like \cref{fig:adv_example} fall into a much closer region compared to testing inputs (e.g., \cref{fig:unseen}), as a result, the truly verifiable region for a given reference input like \cref{fig:anchor} can only be smaller. 
All neural network verification approaches following such data-as-specification paradigm inherit this limitation \emph{regardless} of their underlying verification techniques. 
In order to avoid such a limitation, a new paradigm for specifying what is correct or wrong is necessary.
The intrinsic challenge is that manually giving a proper specification on the input space is no easier than directly 
programming a solution to the machine learning problem itself. 
We envision that a promising way to address this challenge is developing specifications directly on top of, instead of being agnostic to, the learned model.

We propose a new family of specifications, \textit{neural representation as specification}, where neural activation patterns form specifications.
%It is common in software verification that specifications need to be tailored according to program structure and sometimes implementation details. 
The key observation is that inputs from the same class often share a neural activation pattern (NAP) -- a carefully chosen subset of neurons that are expected to be activated (or not activated) for the majority of inputs in a class.
% A dominant NAP captures a very different landscape of inputs beyond a set of balls or boxes centering around reference inputs. 
Although two inputs are distant in a certain norm in the input space, the neural activations exhibited when the same prediction is made are very close.
For instance, we can find a \textit{single} NAP that is shared by \textit{nearly all} training and testing images (including \cref{fig:unseen} and \cref{fig:anchor}) in the same class but not the adversarial example like \cref{fig:adv_example}.
We can further formally \textit{verify} that \textit{all possible} inputs following this particular NAP can never be misclassified. 
Specifications based on NAP enable successful verification of a broad region of inputs, which would not be possible if the data-as-specification paradigm were used. For the MNIST dataset, a verifiable NAP \textit{mined} from the training images could cover up to 84\% testing images, a significant improvement in contrast to 0\% when using neighborhoods of training images as the specification. To our best knowledge, this is the first time that a significant fraction of \textit{unseen} testing images have been formally verified. 

% Interestingly, the verified dominant NAP also enables us to ``double check'' whether ground truth labels given by human beings are indeed reliable. 
% \cref{fig:nap_disagree} shows such an example on which our verified NAP disagrees with the ground truth. 
% % \cref{fig:nap_disagree} might be controversial as it is near the human decision boundary as well. 
% To avoid verifying such a potentially controversial region of inputs, 
% we have a tunable parameter to specialize a dominant NAP. Furthermore, the tunable parameter accounts for the precision-recall trade-off for NAP 

This unique advantage of using NAPs as specification is enabled by the intrinsic information (or neural representation) embedded in the neural network model.
Furthermore, such information is a simple byproduct of a prediction and can be collected easily and efficiently. 
Besides serving as reliable specifications for neural networks, we foresee other important applications of NAPs. 
For instance, verified NAPs may serve as proofs of correctness or certificates for predictions. 
We hope our initial findings shared in this paper would inspire new interesting applications. 
We summarize our contribution as follows:

\begin{itemize}
\item We propose a new family of formal specifications for neural networks, \emph{neural representation as specification}, which use 
 activation patterns (NAPs) as specifications. We also introduce a tunable parameter to specify the level of abstraction of NAPs.

\item We propose a simple yet effective approximate method to mine NAPs from neural networks and training datasets.

\item We show that NAPs can be easily checked by out-of-the-box neural network verification tools used in VNNCOMP -- the annual neural network verification
competition, such as Marabou. 
%encoded into existing neural network verification tools such as Marabou. 
% \ag{Use \emph{spec mininig} or \emph{mining} instead of \emph{extract}. Here, and elsewhere.}

% \item We show NAPs represents fundamental linear regions of the input space corresponding to the distribution of input data through visualization. This explains why NAPs are better candidates than $L_\infty$ norm-balls for neural specifications. 
% \ag{Very weak statement. Restate. Do better than \emph{interesting}}

\item We conduct thorough experimental evaluations from both statistical and formal verification perspectives.
Particularly, we show that a single NAP is sufficient for certifying a significant fraction of unseen inputs. 
% We also identify and verify several crucial properties of NAPs such as no misclassification and no-ambiguity. 
% \ag{Do better than \emph{nice}.}
\end{itemize}

\section{Background}
\label{sec:background}

\subsection{Neural networks for classification tasks}
\label{sec:NN_basic}
In this paper, we focus  on feed-forward neural networks for classification (including convolutional neural nets).
A neural network $\cN$ of $L$ layers is a set $\{(\bW^i, \bb^i) \mid 1 \leq i \leq L\}$, where $\bW^i$ and $\bb^i$ are the weight matrix and the bias for layer $i$, respectively. 
The neural network $\cN$ defined a function $F_{N}: \R^{d_0} \rightarrow \R^{d_L} $ (${d_0}$ and ${d_L}$ represent the input and output dimension, respectively), defined as $F_{\cN}(x) = z^L(x)$,
where $z^0(x) = x$, $z^i(x) = \bW ^i \sigma(z^{i-1}(x)) + \bb^i$
and $\sigma$ is the activation function. Neurons are indexed linearly by $v_0, v_1, \cdots$.
In this paper we focus only on the ReLU activation function, i.e., $\sigma(x) = \max(x, 0)$ element-wise, but the idea and techniques can be generalized for different activation functions and architectures as well. The $i^{th}$ element of the prediction vector $F_{\cN}(x)[i]$ represents the score or likelihood for the $i^{th}$ label, and the one with the highest score ($\arg\max_i F_{\cN}(x)[i]$) is often considered as the predicted label of the network $\cN$. We denote this output label as $\cC_\cN(x)$. When the context is clear, we omit the subscript $\cN$ for simplicity.

% We train XNET on $1\,000$ randomly-generated inputs and achieve the perfect F1-score of 1 on another $1\,000$ random inputs.
% \ag{this ends abruptly. better merge it where it is actually used as an example. This is not a running example at this point since it is not detailed enough.}

\subsection{Adversarial attacks against neural networks and the robustness verification problem}
% \ag{Rewrite this paragraph to separate definition from discussion. First define, then discuss if necessary}
Given a neural network $\cN$, the aim of adversarial attacks is to find a perturbation $\upsilon$ of an input $x$, such that $x$ and $x+\upsilon$ are ``similar'' according to some domain knowledge, yet $\cC(x) \neq \cC(x+\upsilon)$. 
In this paper, we use the common formulation of ``similarity'' in the field: two inputs are similar if the $L_\infty$ norm of $\upsilon$ is small. Under this formulation, finding an adversarial example can be defined as solving the following optimization problem:\footnote{While there are alternative formulations of adversarial robustness (see~\citet{Xu2020AdversarialAA}), in this paper, we use adversarial attacks as a black box, thus, stating one formulation is sufficient.}
\begin{align}
    \min &||\upsilon||_\infty \text{ } 
    s.t. \text{ } \cC(x) \neq \cC(x+\upsilon)
\end{align}

In practice, it is very hard to formally define ``similar'': should an image and a crop of it be ``similar''? Should two sentences differ by one synonym be the same? We refer curious readers to the survey \citep{Xu2020AdversarialAA} for a comprehensive review of different formulations.

One natural defense against adversarial attacks, called \emph{robustness verification}, is to prove that $\min||\upsilon||_\infty $ must be greater than some user-specified threshold $\epsilon$. Formally, given that $\cC(x)=i$, we verify
\begin{align}
\label{eq:robustness_ori}
    \forall x' \in \pertx \cdot \forall j \neq i \cdot F(x')[i] - F(x')[j] > 0
\end{align}
where $\pertx$ is a $L_\infty$ norm-ball of radius $\epsilon$ centered at $x$:
$\pertx = \{x' \mid ||x-x'||_{\infty} \leq \epsilon \}$.
If \cref{eq:robustness_ori} holds, we say that $x$ is $\epsilon$-robust. 

\section{Neural activation patterns}
\label{sec:method}
In this section, we discuss in detail neural activation patterns (NAPs), what we consider as NAPs and how to relax them, and what interesting properties of NAPs can be checked using neural network verification tools like Marabou  \citep{Marabou}.
\subsection{NAPs and their relaxation}
\label{sec:NAP_def}
In our setting (\cref{sec:background}), the output of each neuron is passed to the ReLU function before going to neurons of the next layer, i.e., $z^i(x) = \bW ^i \sigma(z^{i-1}(x)) + \bb^i$. We abstract each neuron into two states: \emph{activated} (if its output is positive) and \emph{deactivated} (if its output is non-positive). Clearly, for any given input, each neuron can be either activated or deactivated.

% \begin{definition}[Neural Activation Pattern] A \emph{Neural Activation Pattern (NAP)} of a neural network $\cN$ is a pair $\cP_\cN \mathbin{:=} (A, D)$, where $A$ and $D$ are two disjoint set of neurons.
% \end{definition}

\begin{definition}[Neural Activation Pattern] A \emph{Neural Activation Pattern (NAP)} of a neural network is a tuple $\cP \mathbin{:=} (A, D)$, where $A$ and $D$ are two disjoint subsets of activated and deactivated neurons, respectively.
\end{definition}

\begin{definition}[Partially ordered NAP] For any given two NAPs $\bar\cP \mathbin{:=} (\bar  A, \bar  D)$ and $\cP \mathbin{:=} (A, D)$. We say $\bar\cP$ subsumes $\cP$ iff $  A,  D$ are subsets of $ \bar A, \bar D$ respectively. Formally, this can be defined as:
\begin{align}
    \bar \cP \preccurlyeq \cP  \text{ } \iff \text{ }  \bar A 	\supseteq A \text{ } and \text{ }  \bar D \supseteq D
\end{align}
Moreover, two NAPs $\bar\cP$ and $\cP$ are equivalent if 
$\bar\cP \preccurlyeq \cP$ and $\cP \preccurlyeq \bar\cP$.
\end{definition}

% \begin{definition}[NAP Comparing Function] A NAP Comparing Function $C$ takes two NAPs $\bar\cP \mathbin{:=} (\bar  A, \bar  D)$ and $\cP \mathbin{:=} (A, D)$  as inputs, and returns a binary output.  More formally, it is defined as:
% \begin{align}
%     C (\bar \cP,\cP) \mathbin{:=} 
%     \begin{cases}
%     1, & \text{if $ \bar A \subseteq A$ and $ \bar D \subseteq D$}.\\
%     0, & \text{otherwise}.
%   \end{cases}
% \end{align}
% \end{definition}
% \allen{partial order}

% \xx{here should we use some other notation for the extracted pattern for $x$ to distinguish it from the specific NAP mentioned later? such as $\cP', A', D'$ or $\cP_x, A_x, D_x$ or something}

\begin{definition}[NAP Extraction Function] A NAP Extraction Function $E$ takes a neural network $\cN$ and an input $x$ as parameters, and returns a NAP $\cP \mathbin{:=} (A, D)$  where $A$ and $D$ represent all the activated and deactivated neurons of $\cN$ respectively when passing $x$ through $\cN$. 
\end{definition}

With the above definitions in mind, we are able to describe the relationship between an input and a specific NAP. An input $x$ \emph{follows} a NAP $\cP$  of a neural network $\cN$ if: 
\begin{align}
    E(\cN,x) \preccurlyeq \cP
\end{align}

For a given neural network $\cN$ and an input $x$, it is possible $x$  follows multiple NAPs. In addition, 
there are some trivial NAPs such as $({\emptyset},{\emptyset})$ that can be followed by any input. From the representational learning point of view, these trivial NAPs are the least specific abstraction of inputs, which fails to represent data with different labels. Thus, we are prone to study more specific NAPs due to their rich representational power. 
Moreover, an ideal yet maybe impractical scenario is that all inputs with a specific label follow the same NAP. Given a label $\ell$, and let $S$ be the training dataset, and $S_\ell$ be the set of data labeled as $\ell$, Formally, this scenario can be described as:
    \begin{align}
        \forall x \in S_\ell  \cdot \text{ } E(\cN,x) \preccurlyeq \cP_{\ell} \iff \cC(x) = \ell
    \end{align}
This can be viewed as a condition for perfectly solving classification problems. In our view, $\cP_{\ell}$, the NAP with respect to $\ell$, if exists, can be seen as a certificate for the prediction of a neural network: inputs following $\cP_{\ell}$ can be provably classified as $\ell$ by $\cN$. However, in most cases, it is infeasible to have a perfect $\cP_{\ell}$ that captures the exact inputs for a given class. On the one hand, there is no access to the ground truth of all possible inputs; on the other hand, DNNs are not guaranteed to precisely learn the ideal patterns. Thus, to accommodate standard classification settings in which Type I and Type II Errors are non-negligible, we relax $\cP_{\ell}$ in such a way that only a portion of the input data with a specific label $\ell$ follows the relaxed NAP. The formal relaxation of NAPs is defined as follows.

% First, instead of quantifying $x$ over the input space, we limit $x$ to a dataset. Second, we break the strong $\iff$ condition, keeping only the condition $\cC(x) = \ell \implies \cP_\ell(x) = True $. Finally, we relax the implication even further, by allowing a neuron to be added to the NAP even when it is not activated/deactivated in all inputs of the same label. We call this relaxed version the $\delta$-relaxed dominant NAP.
% \xs{rewrite}

% \ag{Talk about relaxation of the definition and how useful it might be.}

% \begin{definition}[$\delta$-relaxed dominant NAP] We introduce a relaxing factor $\delta \in [0,1]$. We say a NAP is $\delta$-relaxed dominant with respect to the label $\ell$, denoted as $\cP^\delta_\ell \mathbin{:=} (A^\delta_\ell, D^\delta_\ell) $, if it satisfies the following two conditions:
%     \begin{align}
%         \exists  S'_\ell\subseteq S_\ell \text{ } s.t. \text{ } \frac{|S'_\ell |}{|S_\ell |} \geq \delta \text{ \emph{and} } \forall x \in S'_\ell, \text{ } E(\cN,x) \supseteq A^\delta_\ell \\
%         \exists  S'_\ell\subseteq S_\ell \text{ } s.t. \text{ } \frac{|S'_\ell |}{|S_\ell |} \geq \delta \text{ \emph{and} } \forall x \in S'_\ell, \text{ } E(\cN,x) \supseteq D^\delta_\ell
%     \end{align} 
% \end{definition}

\begin{definition}[$\delta$-relaxed NAP] We introduce a relaxing factor $\delta \in [0,1]$. We say a NAP is $\delta$-relaxed with respect to the label $\ell$, denoted as $\cP^\delta_\ell \mathbin{:=} (A^\delta_\ell, D^\delta_\ell) $, if it satisfies the following condition:
    \begin{align}
        \exists  S'_\ell\subseteq S_\ell \text{ } s.t. \text{ } \frac{|S'_\ell |}{|S_\ell |} \geq \delta \text{ \emph{and} } \forall x \in S'_\ell, \text{ } E(\cN,x) \preccurlyeq \cP^\delta_\ell    \end{align}
    
\end{definition}

% \xs{we need an intuitive explanation in natural language; the follows should be replaced with an construction algorithm}
% \allen{on the relax factor}

% \begin{enumerate}
%     \item Initialize two counters $a_k$ and $d_k$ for each neuron $v_k$. 
%     \item $\forall x \in S_\ell$, compute $F_\cN(x)$, let $a_k\mathrel{+}=1$ if $v_k$ is activated; let $d_k \mathrel{+}= 1$ if $v_k$ is deactivated. 

%     \item $ A_\ell \leftarrow \{ v_k \mid \frac{a_k}{|S_\ell|} \geq 1- \delta \}$ , $D_\ell \leftarrow \{ v_k \mid \frac{d_k}{|S_\ell|} \geq 1- \delta \}$
%     \item $ \cP^\delta_\ell \leftarrow (A_\ell,D_\ell)$
% \end{enumerate}

% \begin{algorithm}[tb]
%    \caption{Bubble Sort}
%    \label{alg:example}
% \begin{algorithmic}
%    \STATE {\bfseries Input:} data $x_i$, size $m$
%    \REPEAT
%    \STATE Initialize $noChange = true$.
%    \FOR{$i=1$ {\bfseries to} $m-1$}
%    \IF{$x_i > x_{i+1}$}
%    \STATE Swap $x_i$ and $x_{i+1}$
%    \STATE $noChange = false$
%    \ENDIF
%    \ENDFOR
%    \UNTIL{$noChange$ is $true$}
% \end{algorithmic}
% \end{algorithm}

\begin{algorithm}[tb]
   \caption{NAP Mining Algorithm}
   \label{alg:NAP Mining Algorithm}
\begin{algorithmic}
    \STATE {\bfseries Input:} relaxing factor $\delta$, neural network $\cN$, dataset $S_\ell$
   \STATE Initialize a counter $c_k$ for each neuron $v_k$
   % \STATE $\forall x \in S_\ell$, compute $E(\cN,x)$.
   \FOR{ $x \in S_\ell$} \STATE compute $E(\cN,x)$
   \IF{$v_k$ is activated}
   \STATE $c_k\mathrel{+}=1$ 
   \ENDIF
   \ENDFOR
   \STATE  $ A_\ell \leftarrow \{ v_k \mid \frac{c_k}{|S_\ell|} \geq \delta \}$, $D_\ell \leftarrow \{ v_k \mid \frac{c_k}{|S_\ell|} \leq 1- \delta \}$ 
   % \STATE $D_\ell \leftarrow \{ v_k \mid \frac{c_k}{|S_\ell|} \leq 1- \delta \}$
   \STATE  $ \cP_\ell^\delta \leftarrow (A_\ell,D_\ell)$
\end{algorithmic}
\end{algorithm}

% \allen{for loop}

% \xx{The above definition is the ideal way of finding the $\delta$-relaxed NAP, however, it was not easy to implement, and our focus of this paper is to show a new potential direction for designing the specification of neural network verification, so, instead, we implemented the \cref{alg:NAP Mining Algorithm} as an approximation to our definition and applied it in our experiments. Also, in our algorithm, there is one thing need to be noticed which is, the value of $\delta$ should be greater than 0.5, otherwise, it will... }
% \NL{Should we move the paragraph to 3.2? Talking about dominant NAP per input here and hinting at dominant NAP per label in the right below section is clunky}
% Furthermore, we call the most precise NAP, i.e. the NAP that contains all activated and deactivated neurons, the dominant NAP \NL{Does that mean the dominant is simply just the set of all neurons?}. This may seem trivial for a single input $x$, as the dominant NAP contains all neurons. However, this is generally not true for dominant NAPs defined with respect to a label, as we will discuss in \cref{sec:method}. Moreover, it is also not hard to show that other NAPs are subsets of the dominant NAP.
% % Vice versa, a single $\NAP_\cN$ can also cover multiple inputs. 
% In short, the dominant NAP allows us to acquire a rich family of NAPs.

Intuitively, the $\delta$-relaxed factor controls the level of abstraction of NAP. When $\delta = 1.0$, not only $\cP^{\delta = 1.0}$ is the most precise (as all inputs from $S_\ell$ follow it) but also the least specific. In this sense, $\cP^{\delta = 1.0}$ can be viewed as the highest level of abstraction of the common neural representation of inputs with a specific label. However, being too abstract is also a sign of under-fitting, this may also enhance the likelihood of Type II Errors for NAPs. By decreasing $\delta$, the likelihood of a neuron being chosen to form a NAP increases, making NAPs more specific. This may help alleviate Type II Errors, yet may also worsen the recall rate by producing more Type I Errors.

In order to effectively mine $\delta$-relaxed NAPs, we propose a simple statistical method shown in Algorithm \ref{alg:NAP Mining Algorithm} \footnote{Note that this algorithm is an approximate method for mining $\delta$-relaxed NAP, whereas $\delta$ should be greater than $0.5$, otherwise, $A^\delta_\ell \bigcap D^\delta_\ell \neq \emptyset$. 
We leave more precise
algorithms for future work.}. \cref{tab:mnist_test_count} reports the effect of $\delta$ on the precision recall trade-off for mined $\delta$-relaxed NAPs on the MNIST dataset. The table shows how many test images from a label $\ell$ follow $\cP^\delta_\ell$, together with how many test images from other labels that also follow the same $\cP^\delta_\ell$.
For example, there are 980 images in the test set with label 0 (second column). Among them, 967 images follow $\cP^{\delta=1.0}_{\ell=0}$. In addition to that, there are 20 images from the other 9 labels that also follow $\cP^{\delta=1.0}_{\ell=0}$. 
\begin{table}[t]
\caption{The number of the test images in MNIST that follow a given $\delta.\NAP$. For a label $i$, $\overline{i}$ represents images with labels other than $i$ yet follow $\delta.\NAP^i$. The leftmost column is the values of $\delta$. The top row indicates how many images in the test set are of a label. } 
\label{tab:mnist_test_count}

\begin{center}
\tiny
\resizebox{\columnwidth}{!}{%
\begin{tabular}{ |l ||r|r|r|r|r|r|r|r|r|r|r|r|r|r|r|r|r|r|r|r| }
\hline

& \multicolumn{2}{r|}{0}& \multicolumn{2}{r|}{1}& \multicolumn{2}{r|}{2}& \multicolumn{2}{r|}{3}& \multicolumn{2}{r|}{4}& \multicolumn{2}{r|}{5}& \multicolumn{2}{r|}{6}& \multicolumn{2}{r|}{7}& \multicolumn{2}{r|}{8}& \multicolumn{2}{r|}{9}\\
& \multicolumn{2}{r|}{(980)}& \multicolumn{2}{r|}{(1135)}& \multicolumn{2}{r|}{(1032)}& \multicolumn{2}{r|}{(1010)}& \multicolumn{2}{r|}{(982)}& \multicolumn{2}{r|}{(892)}& \multicolumn{2}{r|}{(958)}& \multicolumn{2}{r|}{(1028)}& \multicolumn{2}{r|}{(974)}& \multicolumn{2}{r|}{(1009)}\\
\cline{2-21}
&&&&&&&&&&&&&&&&&&&&\\[-0.9em]
&0 & $\overline{0}$ & 1 & $\overline{1}$ & 2 & $\overline{2}$ & 3 & $\overline{3}$ & 4 & $\overline{4}$ & 5 & $\overline{5}$ & 6 & $\overline{6}$ & 7 & $\overline{7}$ & 8 & $\overline{8}$ & 9 & $\overline{9}$\\
\hline\hline
1.00&967 & 20 & 1124 & 8 & 997 & 22 & 980 & 13 & 959 & 25 & 874 & 32 & 937 & 26 & 1003 & 28 & 941 & 22 & 967 & 12\\ 
 \hline
0.99&775 & 1 & 959 & 0 & 792 & 4 & 787 & 2 & 766 & 3 & 677 & 1 & 726 & 4 & 809 & 2 & 696 & 3 & 828 & 4\\ 
 \hline
0.95&376 & 0 & 456 & 0 & 261 & 1 & 320 & 0 & 259 & 0 & 226 & 0 & 200 & 0 & 357 & 0 & 192 & 0 & 277 & 0\\ 
 \hline
0.90&111 & 0 & 126 & 0 & 43 & 0 & 92 & 0 & 76 & 0 & 24 & 0 & 45 & 0 & 144 & 0 & 44 & 0 & 73 & 0\\ 
 \hline
\end{tabular}
}
\end{center}
\end{table}
With the decrease of $\delta$, we can see that in both cases, both numbers decrease, suggesting that it is harder for an image to follow $\cP^{\delta=.99}_{\ell=0}$ without being classified as 0 (the NAP is more precise), at the cost of having many images classified as 0 fail to follow $\cP^{\delta=.99}_{\ell=0}$ (the NAP recalls worse). In short, the usefulness of NAPs largely depends on their precision-recall trade-off. Thus, choosing the right $\delta$ or the right level of abstraction becomes crucial in using NAPs as specifications in verification. We will discuss this matter further in \cref{sec:eval}.

\subsection{Interesting NAP properties}
We expect that NAPs can serve as the key component in more reliable specifications of neural networks. As the first study on this topic, we introduce here three important ones.

\label{sec:NAP_property}
\textbf{The non-ambiguity property of NAPs}
% \NL{Do another pass}
We want our NAPs to give us some confidence about the predicted label of an input, thus a crucial sanity check is to verify that no input can follow two different NAPs of two different labels.
Formally, we want to verify the following:
\begin{align}
    \forall x \cdot \forall i \neq j \cdot E(\cN,x) \preccurlyeq \cP_{\ell=i} \Longrightarrow
    E(\cN,x) \undernegpreccurlyeq \cP_{\ell=j}
\end{align}

Note that this property doesn't hold if either $A_{\ell=i}\bigcap D_{\ell=j}$ or $A_{\ell=j} \bigcap D_{\ell=i}$ is non-empty as a single input cannot activate and deactivate the same neuron. If that's not the case, we can encode and verify the property using verification tools.
% \xs{polish the wording; trivial does not sound good, also its meaning (true or false) is not clear. }
%\cref{alg:check_definitive}
% in \cref{Appendix}.

% \begin{align}
%     \forall x \cdot \forall i, j (i\neq j) \cdot \cP_{\ell=i}(x) = True \implies \cP_{\ell=j}(x) = False
% \end{align}

\textbf{NAP robustness property} 
The intuition of using neural representation as specification not only accounts for the internal decision-making process of neural networks but also leverages the fact that NAPs themselves map to regions of our interests in the whole input spaces. In contrast to canonical $\epsilon$-balls, these NAP-derived regions are more flexible in terms of size and shape. We explain this insight in more detail in \cref{sec:case_study}.
Concretely, we formalize this NAP robustness verification problem as follows: given a neural network $\cN$ and a NAP $\cP_{\ell=i}$, we want to check:
% \xx{For $R$. should we make it to be $R(\cP_{\ell=i})$? to make it align with the $B^+$'s definition later}
\begin{align}
    \forall x \in R \cdot \forall j \neq i \cdot F(x)[i] - F(x)[j] > 0
\end{align}
in which 
\begin{align}
    R = \{ x \mid {E(\cN,x) \preccurlyeq \cP_{\ell=i}}  \}  
\end{align}

\textbf{NAP-augmented robustness property} Instead of only having the activation patterns as specification, we can still specify {$\epsilon$-balls in the input space for verification. This conjugated form of specification has two advantages: First, it focuses on the verification of valid test inputs instead of adversarial examples. Second, the constraints on NAPs are likely to make verification tasks effortless by refining the search space of the original verification problem, in most cases, allowing the verification on much larger $\epsilon$-balls. We formalize the NAP-augmented robustness verification problem as follows: given a neural network $\cN$, an input $x$, and a mined $\cP_{\ell=i}$, we check:
\begin{align}
    \forall x' \in B^{+}(x, \epsilon, \cP_{\ell=i}) \cdot \forall j \neq i \cdot F(x')[i] - F(x')[j] > 0
\end{align}
in which $\cC(x) = i$ and
\begin{align}
    B^{+}(x, \epsilon, \cP_{\ell=i}) = \{ x' \mid ||x-x'||_\infty \leq \epsilon , {E(\cN,x') \preccurlyeq \cP_{\ell=i}} \}
\end{align}

\textbf{Working with NAPs using Marabou} In this paper, we use Marabou \citep{Marabou}, a dedicated state-of-the-art NN verifier. Marabou extends the Simplex \cite{simplex} algorithm for solving linear programming with special mechanisms to handle non-linear activation functions. Internally, Marabou encodes both the verification problem and the adversarial attacks as a system of linear constraints (the weighted sum and the properties) and non-linear constraints (the activation functions). Same as Simplex, at each iteration, Marabou tries to fix a variable so that it doesn't violate its constraints. While in Simplex, a violation can only happen due to a variable becoming out-of-bound, in Marabou a violation can also happen when a variable doesn't satisfy its activation constraints.

NAPs and NAP properties can be encoded using Marabou with little to no changes to Marabou itself. To force a neuron to be activated or deactivated, we add a constraint for its output. To improve performance, we infer ReLU's phases implied by the NAPs, and change the corresponding constraints\footnote{Marabou has a similar optimization, but the user cannot control when or if it is applied.}.For example, given a ReLU $v_i=max(v_k, 0)$, to enforce $v_k$ to be activated, we remove the constraint from Marabou and add two new ones: $v_i = v_k$, and $v_k \geq 0$.

\subsection{Case Study: Visualizing NAPs of a simple Neural Network}
\label{sec:case_study}
We show the advantages of NAPs as specifications using a simple example of a three-layer feed-forward neural network that predicts a class of 20 points located on a 2D plane. We trained a neural network consisting of six neurons that achieves 100\% accuracy in the prediction task. The resulting linear regions as well as the training data are illustrated in \cref{fig:linear_regions}. \Cref{tab:case_study} summarizes the frequency of states of each neuron based on the result of passing all input data through the network, and NAPs for labels 0 and 1. 
% It is also not hard to see that NAPs are abstractions of neuron states. 
% \ag{Replace \emph{as we can see} with something more direct. You want reader to see, not us}

\cref{fig:NAP_visual} visualizes NAPs for labels 0 and 1, and the unspecified region which provides no guarantees on data that fall into it. The green dot is so close to the boundary between $\cP_{\ell=0}$ and the unspecified region that some $L_\infty$ norm-balls (boxes) such as the one drawn in the dashed line may contain an adversarial example from the unspecified region. Thus, what we could verify ends up being a small box within $\cP_{\ell=0}$. However, using $\cP_{\ell=0}$ as a specification allows us to verify a much more flexible region than just boxes, as suggested by the NAP-augmented robustness property in Section \ref{sec:NAP_property}. This idea generalizes beyond the simple 2D case, and we will illustrate its effectiveness further with a critical evaluation in Section \ref{sec:NAP_aug}.
\begin{figure}[t]
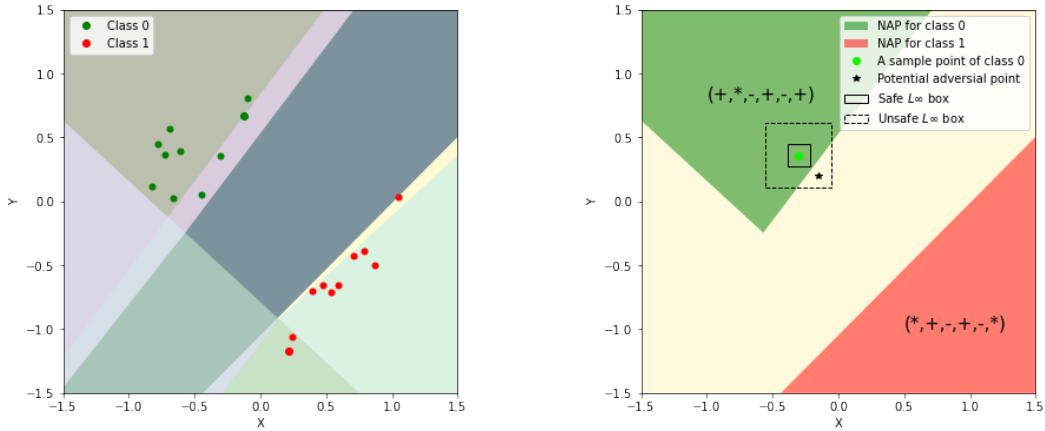

     \begin{subfigure}[t]{0.45\textwidth}
         \centering         \includegraphics[width=\textwidth]{figures/originalNAP.pdf}
         \caption{Linear regions in different colors are determined by weights and biases of the neural network. Points colored either red or green constitute the training set.} 
         \label{fig:linear_regions}
     \end{subfigure}
     \hfill
     \begin{subfigure}[t]{0.45\textwidth}
         \centering         \includegraphics[width=\textwidth]{figures/adversial_point.pdf}
         \caption{NAPs are more flexible than $L_\infty$ norm-balls (boxes) in terms of covering verifiable regions.}
         \label{fig:NAP_visual}
         \hfill
     \end{subfigure}
        \caption{Visualization of linear regions and NAPs as specifications compared to $L_\infty$ norm-balls.}
        
        \label{fig:nap reg}

\end{figure}

\begin{table}[t]
\caption{The frequency of each ReLU and the dominant NAPs for each label. Activated and deactivated neurons are denoted by $+$ and $-$, respectively, and $*$ denotes an arbitrary neuron state.} 
\label{tab:case_study}
\centering
\begin{tabular}{ l | r | c| r}
\bf Label & \bf Neuron states & \bf \#samples & \bf Dominant NAP \\
\hline
\hline
\multirow{2}{*}{0 (Green)}&$(+,-,-,+,-,+)$&8& \multirow{2}{*}{$(+,*,-,+,-,+)$} \\ 
 %\cline{2-3}
&$(+,+,-,+,-,+)$&2&  \\ 
 \hline
\multirow{3}{*}{1 (Red)}&$(+,+,-,-,+,-)$&7&  \multirow{3}{*}{$(*,+,-,+,-,*)$}\\ 
%\cline{2-3}
&$(-,+,-,-,+,-)$&2& \\ 
% \cline{2-3}
&$(+,+,-,-,+,+)$&1& \\ 
 \hline

\end{tabular}
\end{table}

\section{Evaluation}
In this section, we validate our observation about the distance between inputs, as well as evaluate our NAPs and NAP properties on networks and datasets from VNNCOMP-21.
\label{sec:eval}
\subsection{Experiment setup}
Our experiments are based on benchmarks from \citet{vnncomp2021} -- the annual neural network verification competition. We use 2 of the datasets from the competition: MNIST and CIFAR10. For MNIST, we use the two largest models \texttt{mnistfc\_256x4} and \texttt{mnistfc\_256x6}, a 4- and 6-layers fully connected network with 256 neurons for each layer, respectively. For CIFAR10, we use the convolutional neural net \texttt{cifar10\_small.onnx} with 2568 ReLUs. 
Experiments are done on a machine with an Intel(R) Xeon(R) CPU E5-2686 and 480GBs of RAM. Timeouts for MNIST and CIFAR10 are 10 and 30 minutes, respectively.

\subsection{$L_2$ and $L_\infty$ maximum verified bounds}
\label{sec:max_bound}
We empirically find that the $L_2$ and $L_\infty$ maximum verifiable bounds are much smaller than the distance between real data, as illustrated in Figure \ref{fig:failure of distance}. We plot the distribution of distances in $L_2$ \footnote{The $L_2$ metric is not 
 commonly used by the neural network verification research community as it is less computationally efficient than the $L_\infty$  metric. } and $L_\infty$ norm between all pairs of images with the same label from the MNIST dataset, as shown in Figure \ref{fig:mnist_norm}. 
% \ag{What are \emph{maximum norm-based verifiable bounds}. I know what you mean by I don't think it has been defined}
For each class, the smallest $L_\infty$ distance of any two images is significantly larger than 0.05, which is  the largest perturbation used
in \citet{vnncomp2021}. 

\begin{figure}[t]
     \centering
     % \begin{subfigure}[t]{0.32\textwidth}
     %     \centering        \includegraphics[width=\textwidth]{figures/mnist_L1.png}
     %     \caption{The distribution of $L_1$-norms of all image pairs for each label.}
     %     \label{fig:mnist_L1}
     % \end{subfigure} 
     % \hfill
     \begin{subfigure}[t]{0.48\textwidth}
         \centering
         \includegraphics[width=0.8\textwidth]{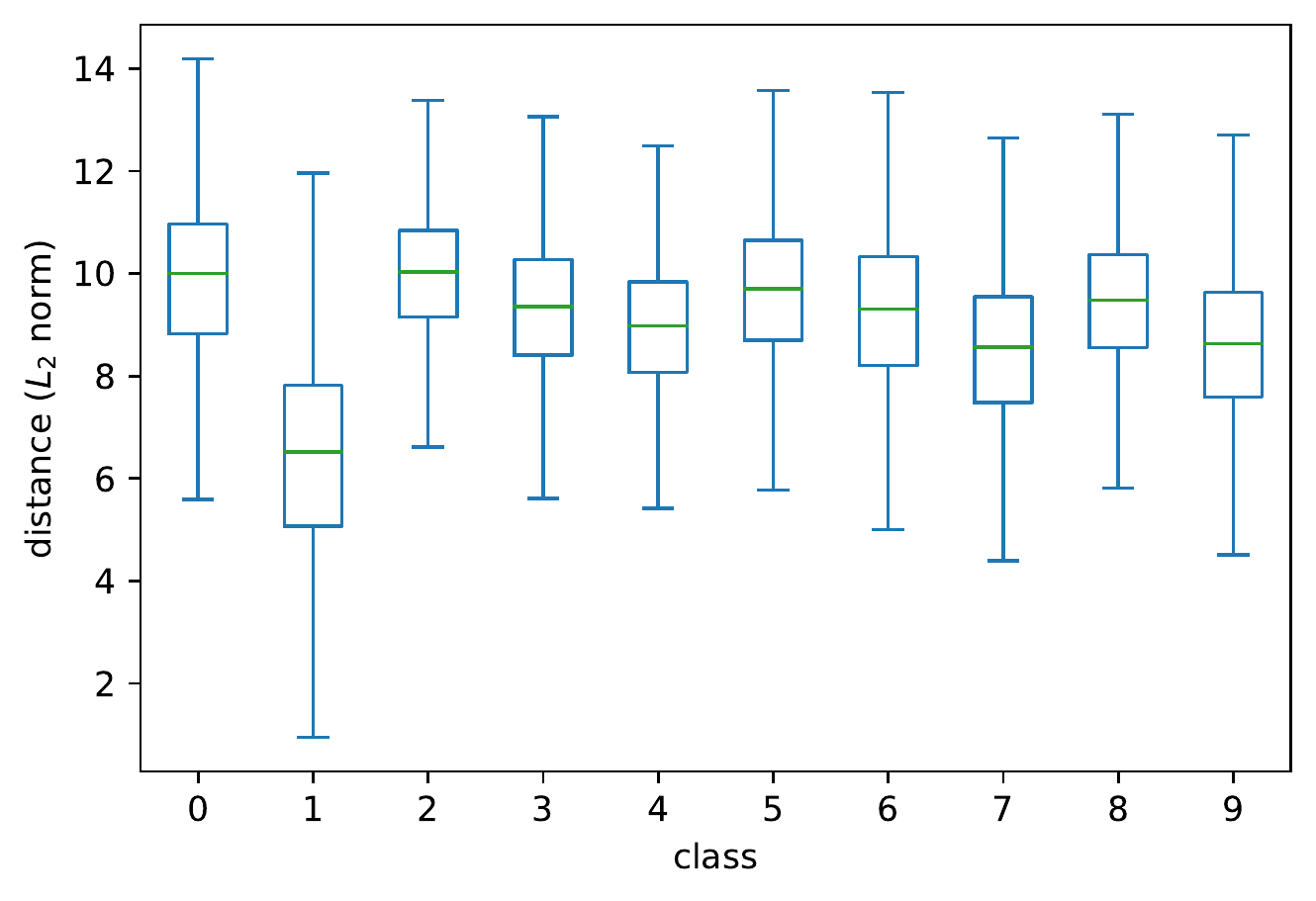}
         \caption{The distribution of $L_2$-norms between any two images from the same label. Images of digit (label) 1 are much similar than that of other digits.}
         \label{fig:mnist_L2}
     \end{subfigure}
     \hfill
     \begin{subfigure}[t]{0.51\textwidth}
         \centering         
         \includegraphics[width=0.8\textwidth]{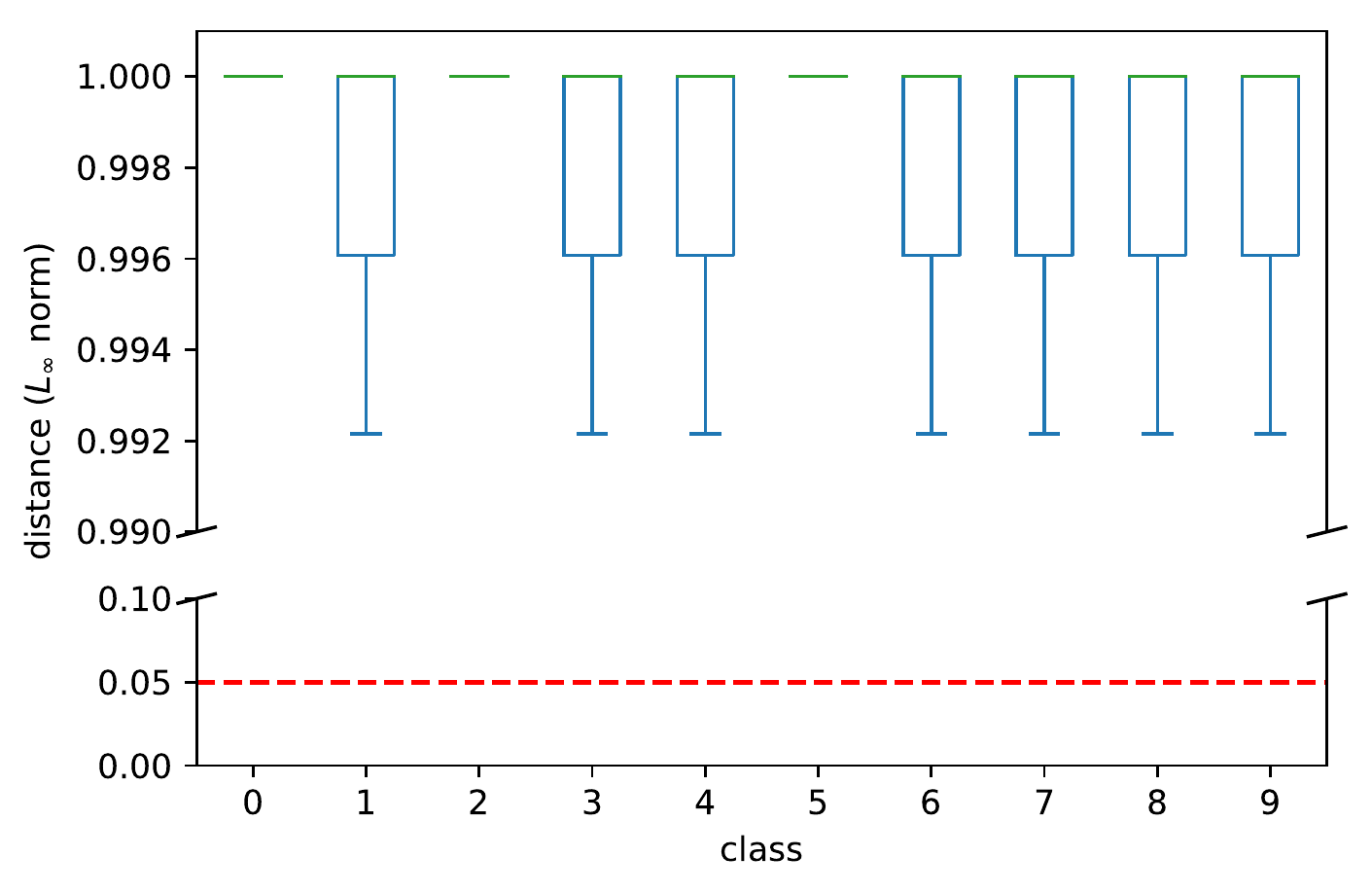}
         \caption{The distribution of $L_\infty$-norms between any two images from the same label. The red line is drawing at 0.05 -- the largest $\epsilon$ used in \citet{vnncomp2021}.}
         \label{fig:mnist_Linf}
     \end{subfigure}
        \caption{Distances between any two images from the same label (class) are quite significant under different metrics of norm.}
        
        \label{fig:mnist_norm}
\end{figure}
This suggests that the \textit{data as specification} paradigm (i.e., using reference inputs with perturbations bounded in $L_2$ or $L_\infty$ norm) is not sufficient to verify test set inputs or unseen data. The differences between training and testing data of each class are usually significantly larger than the perturbations allowed in specifications using $L_\infty$ norm-balls.

\subsection{The NAP robustness property}
\label{sec:NAP_aug}
We conduct two sets of experiments with MNIST and CIFAR10 to demonstrate the NAP and NAP-augmented robustness properties. The results are reported in \cref{tab:fig_1_with_NAP,tab:mnist_augmented_NAP,tab:cifar10_augmented_NAP}.
%\cref{tab:fig_1_with_NAP}, \ cref{tab:mnist_augmented_NAP}, and \cref{tab:cifar10_augmented_NAP}. 
For label $\ell$ from 0 to 9, `Y' (`N') indicates that the network is (not) robust (i.e., no adversarial example of label $\ell$ exists). `T/o' means the verification of robustness timed out.
% , `Cr' means Marabou runs into a numerical instability issue and crashes.   

% In both result tables, for a column under $\ell$, `Y' means no adversarial example exists -- the input is robust against the label $\ell$, `N' means an adversarial example is found -- the input is not robust, and `T/o' means timeout.

\begin{table}[t]
\caption{Robustness of the illustrative example in \Cref{fig:failure of distance}}
\vskip 0.1in
\begin{center}
\label{tab:fig_1_with_NAP}
\resizebox{\columnwidth}{!}{%
\begin{tabular}{ |l | r r r r r r r r r r| }
\hline
& 0& 1& 2& 3& 4& 5& 6& 7& 8&9\\
\hline\hline
$ \epsilon = 0.2$, no NAP &N &- &N &T/o &T/o &T/o &T/o &T/o &N &T/o\\ 
\hline
 $\epsilon = 0.2, \delta = 1.0$ &N &- &N &\Y &T/o &T/o &\Y &T/o &N &N\\ 
 \hline
$ \epsilon = 0.2, \delta = 0.99$ &\Y &- &\Y &\Y &\Y &\Y &\Y &\Y &\Y &\Y\\ 
\hline
  % $\epsilon = 1.0, \delta = 0.99$ &\Y & - &\Y&\Y &\Y &\Y &\Y &\Y &\Y &\Y\\ 
no $\epsilon, \delta = 0.99$ & \multirow{2}{*}{\Y}&\multirow{2}{*}{-}  &\multirow{2}{*}{\Y}&\multirow{2}{*}{\Y}&\multirow{2}{*}{\Y}&\multirow{2}{*}{\Y} &\multirow{2}{*}{\Y}&\multirow{2}{*}{\Y}&\multirow{2}{*}{\Y}&\multirow{2}{*}{\Y}\\ 
   % $\epsilon = 1.0, \delta = 0.99$ &\multirow{2}{*}{\Y}&&&&&&&&&\\
\text{(NAP robustness property)} &&&&&&&&&&\\ 

 \hline
\end{tabular}
}
\end{center}
\end{table}

% \begin{table*}[t]
% \caption{Inputs that are not robust can be augmented with a NAP to be robust. We index the 6 known non-robust inputs from MNIST $x_0$ to $x_5$, and augment them with NAP. }
% \label{tab:mnist_augmented_NAP}
% \vskip 0.1in
% \begin{center}
% \begin{tabular}{ |l |l || r r r r r r r r r r| }
% \hline
% && 0& 1& 2& 3& 4& 5& 6& 7& 8&9\\
% \hline\hline
% $\cC(x_0) = 0$&$ \epsilon = 0.03, \delta = 0$&- &\Y &\Y &\Y &\Y &\Y &\Y &\Y &\Y &\Y\\ 
%  &$\epsilon = 0.3, \delta = 0.01$&- &\Y &\Y &\Y &\Y &\Y &\Y &\Y &\Y &\Y\\ 
%  \hline
% $\cC(x_1) = 1$&$ \epsilon = 0.05, \delta = 0$ &\Y &- &\Y &\Y &\Y &\Y &\Y &\Y &N &\Y\\ 
%  &$\epsilon = 0.3, \delta = 0.01$ &\Y &- &\Y &\Y &\Y &\Y &\Y &\Y &\Y &\Y\\ 
%  \hline
%  $\cC(x_2) = 0$&$ \epsilon = 0.05, \delta = 0$ &- &T/o &T/o &\Y &T/o &T/o &\Y &N &T/o &T/o\\ 
%   &$\epsilon = 0.3, \delta = 0.01$ &- &\Y &\Y &\Y &\Y &\Y &\Y &\Y &\Y &\Y\\
%  \hline
%  $\cC(x_3) = 7$&$ \epsilon = 0.05, \delta = 0$ &N &T/o &\Y &\Y &T/o &T/o &\Y &- &N &T/o\\ 
%   &$\epsilon = 0.3, \delta = 0.01$ &\Y &\Y &\Y &\Y &\Y &\Y &\Y &- &\Y &\Y\\
%  \hline
% $\cC(x_4) = 9$&$ \epsilon = 0.05, \delta = 0$ &T/o &\Y &\Y &\Y &\Y &\Y &N &\Y &N &-\\ 
%  &$\epsilon = 0.3, \delta = 0.01$ &\Y &T/o &T/o &\Y &N &\Y &T/o &T/o &T/o &-\\ 
%  \hline
% $\cC(x_5) = 1$&$ \epsilon = 0.05, \delta = 0$ &\Y &- &N &\Y &\Y &\Y &\Y &N &N &N\\ 
%   &$\epsilon = 0.3, \delta = 0.01$ &\Y & - &\Y &\Y &\Y &\Y &\Y &\Y &\Y &\Y\\ 
%  \hline
% \end{tabular}
% \end{center}
% \end{table*}
\begin{table}[t]
\caption{Inputs that are not robust can be augmented with a NAP to be robust. With $\delta = 0.99$, all inputs can be verified to be robust at $\epsilon=0.05$ -- the largest checked $\epsilon$ in VNNCOMP-21(not shown)
% We extract and index all known non-robust inputs from MNIST $x_0$ to $x_5$, and augment them with NAP. 
}
\label{tab:mnist_augmented_NAP}
\vskip 0.1in
\begin{center}
% \resizebox{\columnwidth}{!}{%
\begin{tabular}{  |l || r r r r r r r r r r| }
\hline
& 0& 1& 2& 3& 4& 5& 6& 7& 8&9\\
\hline\hline
$\mathbf{\cC(x_0) = 0}$ & &&&&&&&&&\\
$ \epsilon = 0.05, \delta = 1.0$&- &\Y &\Y &\Y &\Y &\Y &\Y &\Y &\Y &\Y\\ 
% $ \epsilon = 0.05, \delta = 0.99$&- & & & & & & & & & \\ 
 $\epsilon = 0.3, \delta = 0.99$&- &\Y &\Y &\Y &\Y &\Y &\Y &\Y &\Y &\Y\\ 
 \hline
 $\mathbf{\cC(x_1) = 1}$& &&&&&&&&& \\
$ \epsilon = 0.05, \delta = 1.0$ &\Y &- &\Y &\Y &\Y &\Y &\Y &\Y &N &\Y\\ 
 $\epsilon = 0.3, \delta = 0.99$ &\Y &- &\Y &\Y &\Y &\Y &\Y &\Y &\Y &\Y\\ 
 \hline
  $\mathbf{\cC(x_2) = 0}$& &&&&&&&&&\\
$ \epsilon = 0.05, \delta = 1.0$ &- &T/o &T/o &\Y &T/o &T/o &\Y &N &T/o &T/o\\ 
  $\epsilon = 0.3, \delta = 0.99$ &- &\Y &\Y &\Y &\Y &\Y &\Y &\Y &\Y &\Y\\
 \hline
  $\mathbf{\cC(x_3) = 7}$& &&&&&&&&&\\
$ \epsilon = 0.05, \delta = 1.0$ &N &T/o &\Y &\Y &T/o &T/o &\Y &- &N &T/o\\ 
  $\epsilon = 0.3, \delta = 0.99$ &\Y &\Y &\Y &\Y &\Y &\Y &\Y &- &\Y &\Y\\
 \hline
 $\mathbf{\cC(x_4) = 9}$& &&&&&&&&&\\
$ \epsilon = 0.05, \delta = 1.0$ &T/o &\Y &\Y &\Y &\Y &\Y &N &\Y &N &-\\ 
 $\epsilon = 0.3, \delta = 0.99$ &\Y &T/o &T/o &\Y &N &\Y &T/o &T/o &T/o &-\\ 
 \hline
 $\mathbf{\cC(x_5) = 1}$& &&&&&&&&&\\
$ \epsilon = 0.05, \delta = 1.0$ &\Y &- &N &\Y &\Y &\Y &\Y &N &N &N\\ 
  $\epsilon = 0.3, \delta = 0.99$ &\Y & - &\Y &\Y &\Y &\Y &\Y &\Y &\Y &\Y\\ 
  \hline
  $\mathbf{\cC(x_6) = 9}$& &&&&&&&&&\\
$ \epsilon = 0.05, \delta = 1.0$ &T/o &T/o &T/o &T/o &T/o &T/o &T/o &T/o &T/o &-\\ 
  $\epsilon = 0.3, \delta = 0.99$ &\Y &\Y &\Y &\Y &\Y &\Y &\Y &\Y &\Y &-\\ 
  (\texttt{mnistfc\_256x6})& &&&&&&&&&\\
 \hline
\end{tabular}
% }
\end{center}
\end{table}

\textbf{MNIST with fully connected NNs} In \Cref{fig:failure of distance}, we show an illustrative image $\cI$ (of digit 1) and its adversarial example within the distance of $L_\infty = 0.2$. As shown in \cref{tab:fig_1_with_NAP}, three different kinds of counter-example can be found within this distance.
In contrast, the last row shows that all input images in the \emph{entire input space} following the mined NAP specification 
$\cP^{\delta=0.99}_{\ell=1}$ can be safely verified. 
It is worth noting that this specification covers 84\%  (959/1135) of the test set inputs (\cref{tab:mnist_test_count}). To our best knowledge, this is the first specification for MNIST dataset that covers a substantial fraction of testing images. To some extent, it serves as a candidate machine-checkable definition of digit 1 in MNIST. 

The second and third rows of \cref{tab:fig_1_with_NAP} show the robustness of combining 
  $L_\infty$ norm perturbation and NAPs as the specification. The third row is well expected as the last row has shown that the network is robust against NAP itself (without $L_\infty$ norm constraint). It is interesting to see that when we increase $\delta$ to 1.0, the mined NAP specification $\cP^{\delta=1.0}_{\ell=1}$ becomes too general and covers a much larger region that includes more than 99\% (1124/1135) testing images as shown in \cref{tab:mnist_test_count}. As a result, together with $L_\infty=0.2$ constraint, only two classes of adversarial examples can be safely verified, 
 which is still better than only using $L_\infty=0.2$ perturbation as the specification. 

% While using $\cP^{\delta=1.0}$ as an additional specification, we are able to verify the robustness of $\cI$ against 2 other labels. However, if we decrease $\delta$ to 0.99, we are able to verify the robustness against \emph{all} other labels. More surprisingly, the NAP robustness property is also verified with $\delta=0.99$ (denoted by $\epsilon=1.0$ in the last row). Given that 84\% of the test set inputs also follows $\cP^{\delta=.99}_{\ell=1}$ (\cref{tab:mnist_test_count}), this shows a significant advantage of NAP for verifying inputs sampled from the underlying data distributions compared to traditional specifications. In addition, to the best of our knowledge, this is the first time a significant region of the input space can be formally verified.

% \allen{change the pitch, positive, before all adv. , now better}
% Although we can prove the robustness of the instance $\mathbf{\cC(x_0) = 0}$ using the augmented specifications $B^{+}(\cdot, \epsilon = 0.05, \cP^{\delta = 1.0})$, there are adversarial examples still exist against certain labels, suggesting we need account for the precision-recall trade-off. 

We further study how NAP-augmented specification helps to improve the verifiable bound. 
Specifically, we collect all $(x,\epsilon)$ tuples in VNNCOMP-21 MNIST benchmarks that are known to be not robust (an adv. example is found in $B(x, \epsilon)$). Among them, the first six tuples correspond to \texttt{mnistfc\_256x4} and the last one corresponds to \texttt{mnistfc\_256x6}. \cref{tab:mnist_augmented_NAP} reports the verification results with NAP augmented specification. 

For the first six instances, using the NAP augmented specifications $B^{+}(\cdot, \epsilon = 0.05, \cP^{\delta = 1.0})$ enables the verification against more labels, outperforming using only $L_\infty$ perturbation as the specification. By slightly relaxing the NAP ($\delta=0.99$) , \emph{all} of the chosen inputs can be proven to be robust. Furthermore, with $\delta=0.99$, we can verify the robustness for 6 of the 7 inputs (\cref{tab:mnist_augmented_NAP}) with $\epsilon=0.3$, which is \emph{an order of magnitude} bigger bound than before. Note that decreasing $\delta$ specifies a smaller region, usually allowing verification with bigger $\epsilon$, but a smaller region tends to cover fewer testing inputs. Thus, choosing an appropriate $\delta$ is crucial for having useful NAPs.

% (i.e., specifying a smaller region)
% Unfortunately, we are not able to verify bigger $\epsilon$, except for $x_1$ and $x_5$, of which the NAP
% robustness property can be verified, as demonstrated with the previous experiment. One might tempt to decrease $\delta$ even further to verify bigger $\epsilon$, but one must remember that comes with sacrificing the recall of the NAP. Thus, choosing an appropriate $\delta$ is crucial for having useful NAPs.

% \allen{nham, I comment out the above paragraph, it sounds a little negative, and provides less info, what do you think?}

\textbf{CIFAR10 with CNN} To show that our insights and methods can be applied to more complicated datasets and network topologies, we conduct the second set of experiments using convolutional neural nets trained on the CIFAR10 dataset. We extract all $(x, \epsilon)$ tuples in the CIFAR10 dataset that are known to be not robust from VNNCOMP-21 (an adv. example is found in $B(x, \epsilon)$) and verify them using augmented NAP. For CIFAR10, $\cP^{\delta=1.0}$ does not exist, thus we use $\cP^{\delta=.99}$, $\cP^{\delta=.95}$ and $\cP^{\delta=.90}$. 
%For the input that does not follow the NAP exactly, we skip fixing the ReLUs those are not matched. 
We follow the scenario used in VNNCOMP-21 and test the robustness against  $(\textit{correctLabel} + 1) \mod 10$. The results are reported in \cref{tab:cifar10_augmented_NAP}. As with MNIST, we observe that by relaxing $\delta$, we were able to verify more examples at every $\epsilon$. Even with $\epsilon=0.12$ ($10\times$ the verifiable bound, which translates to an input space $10^{3072}\times$ bigger!), by slightly relaxing $\delta$ to 0.9, we can verify 3 out of 7 inputs.

\subsection{The non-ambiguity property of mined NAPs}
We evaluate the non-ambiguity property of our mined NAP at different $\delta$s on MNIST. At $\delta = 1.0$, we can construct inputs that follow any pair of NAP, indicating that  $\cP^{\delta=1.0}$s do not satisfy the property. However, by setting $\delta=0.99$, we are able to prove the non-ambiguity for \emph{all} pairs of NAPs, through both trivial cases and invoking Marabou. This is because relaxing $\delta$ leaves more neurons in NAPs, making it more difficult to violate the non-ambiguity property.

\begin{table}
\caption{Augmented robustness with CIFAR10 and CNN.}
\label{tab:cifar10_augmented_NAP}
\begin{center}
% \resizebox{12cm}{!}{%
\begin{tabular}{|l|c|c|c|c|c|c|c|c|c|}
\hline
\multicolumn{1}{|c|}{$\epsilon$} & \multicolumn{3}{c|}{0.012} & \multicolumn{3}{c|}{0.024} & \multicolumn{3}{c|}{0.12}\\
\cline{2-10}
\multicolumn{1}{|c|}{$\delta$} & 0.99 & 0.95 & 0.9 & 0.99 & 0.95 & 0.9 & 0.99 & 0.95 & 0.9\\
\hline
$\mathbf{\cC(x_0)=8}$   &\Y &\Y &\Y &N  &T/o  &\Y &T/o &\Y &\Y\\
$\mathbf{\cC(x_1)=6}$   &T/o &N &\Y &N  &N  &\Y &N &N &\Y\\
$\mathbf{\cC(x_2)=0}$   &\Y &\Y &\Y &\Y  &\Y  &\Y &N &N &N\\
$\mathbf{\cC(x_3)=1}$   &N &N &N &N  &N  &N &N &N &N\\
$\mathbf{\cC(x_4)=9}$   &N &\Y &\Y &N  &N  &N &N &N &N\\
$\mathbf{\cC(x_5)=7}$   &\Y &\Y &\Y &N  &T/o  &\Y &N &\Y &\Y\\
$\mathbf{\cC(x_6)=3}$   &\Y &\Y &\Y &\Y  &\Y  &\Y &N &N &N\\
\hline
\end{tabular}
% }
\end{center}
\end{table}

% Like the previous experiment, we observe the benefit of relaxing $\delta$ here: the most precise NAPs contain so few ReLUs that it is very easy to violate the non-ambiguity property, thus relaxing $\delta$ is crucial for NAPs to be useful.

% On the other hand, the \NL{What is the other hand of this `the other hand`?}
The non-ambiguity property of NAPs holds an important prerequisite for neural networks to achieve a sound classification result. Otherwise, the final prediction of inputs with two different labels may become indistinguishable. We argue that mined NAPs should demonstrate strong non-ambiguity properties and ideally, all inputs with the same label $i$ should follow the same $\cP_{\ell=i}$. However, this strong statement may fail even for an accurate model when the training dataset itself is problematic, as what we observed in \cref{fig:nap_disagree} as well as many examples in \cref{appendix:misclassification}. These examples are not only similar to the model but also to humans despite being labeled differently. The experiential results also suggest our mined NAPs do satisfy the strong statement proposed above if excluding these noisy samples.

% We want to point out the connection with \cref{tab:mnist_test_count}: at $\delta=0$, for a given label, i.e 0, there are a few images $(20/9,020)$ from non-0 labels that follow $0.\cP^0$. That is confirmed by checking the non-ambiguity property. However, we also see a few images from non-0 labels that follow $0.01.\cP^0$. How can it be possible? Upon closer inspection, we learn that those inputs actually do not follow any $0.01\cP$, even the one for its own label! Thus, they do not trigger 2 checked NAPs at the same time, escaping our check.

% To conclude, this section presents a study that verifies our hypothesis about $L$-norm distances between data points in the MNIST dataset, thus explaining the limitation of using $L$-norm in verifying neural networks' robustness. Then, we conduct experiments to show that our $\delta$-relaxed dominant NAPs can be used to verify a much bigger region in the input space, which has the potential to generalize to real-world scenarios.

% as well as be used to 

\section{Related work  and Future Directions}
\label{sec:related}
\textbf{Abstract Interpretation in verifying Neural Networks} The software verification problem is undecidable in general \citep{rice_theorem}. Given that a Neural Network can also be considered a program, verifying any non-trivial property of a Neural network is also undecidable. Prior work on neural network verification includes specifications that are linear functions of the output of the network: 
Abstract Interpretation (AbsInt) \citep{Absint} pioneered a happy middle ground: by sacrificing completeness, an AbsInt verifier can find proof much quicker, by over-approximating reachable states of the program. Many NN-verifiers have adopted the same technique, such as DeepPoly \cite{deeppoly}, CROWN \cite{abc}, NNV \cite{nnv}, etc. They all share the same insight: the biggest bottle neck in verifying Neural Networks is the non-linear activation functions. By abstracting the activation into linear functions as much as possible, the verification can be many orders of magnitude faster than complete methods such as Marabou. However, there is no free lunch: Abstract-based verifiers are inconclusive and may not be able to verify properties even when they are correct.\footnote{Methods such as alpha-beta CROWN \cite{abc} claim to be complete even when they are Abstract-based because the abstraction can be controlled to be as precise as the original activation function, thus reducing the method back to a complete one.} On the other hand, the \textit{neural representation as specification} paradigm proposed in this work can be naturally viewed as a method of Abstract Interpretation, in which we abstract the state of each neuron to only activated and deactivated by leveraging NAPs. We would like to explore more refined abstractions such as $\{(-\infty],(0,1],(1,\infty]\}$ in future work. 

\textbf{Neural Activation Pattern in interpreting Neural Networks} 
% \NL{Please keep this part sort. We are already over the page limit.}
% Explaining why a neural network  comes up with a prediction has been a challenging problem, especially as models get increasingly complicated. 
% \allen{I drop this sentence. Now we can fit it in 9 pages}
There are many attempts aimed to address the black-box nature of neural networks by highlighting important features in the input, such as Saliency Maps \cite{saliency_map, grad_cam} and LIME \cite{lime}. But these methods still pose the question of whether the prediction and explanation can be trusted or even verified. Another direction is to consider the internal decision-making process of neural networks such as Neural Activation Patterns (NAP). One popular line of research relating to NAPs is to leverage them in feature visualization  \cite{vis_network, NAPs1, Erhan2009VisualizingHF}, which investigates what kind of input images could activate certain neurons in the model. 
% \NL{Please keep this part sort. We are already over the page limit. :'(}
Those methods also have the ability to visualize the internal working mechanism of the model to help with transparency. This line of methods is known as activation maximization. While being great at explaining the prediction of a given input, activation maximization methods do not provide a specification based on the activation pattern: at best they can establish a correlation between seeing a pattern and observing an output, but not causality. Moreover, moving from reference sample to revealing neural network activation pattern is limiting as the portion of NAP uncovered is dependent on the input data. This means that it might not be able to handle cases of unexpected test data. Conversely, our method starts from the bottom up: from the activation pattern, we uncover what region of input can be verified. This property of our method grants the capability to be generalized. Motivated by our promising results, we would like to generalize our approach to modern deep learning models such as Transformers \cite{transformer}, which employ much more complex network structures than a simple feed-forward structure. Gopinath et al. also focus on neural network explanation by studying input properties and layer properties  \cite{DBLP:conf/kbse/GopinathCPT19}. The latter property is the most related to our work. They collect activation of all neurons in a specific layer and then learn a decision tree by viewing each activation as a feature. The learned decision tree can capture many different activation patterns in the concerned layer and can serve as a formal (logical) interpretation for that particular layer. In contrast, we aim to find a dominant neural activation pattern (across layers) and further verify that it captures a large fraction of desired inputs from a specific class and meanwhile does not contain any adversarial input.

\section{Conclusion}
\label{sec:conclusion}

We propose a new paradigm of neural network specifications, which we call \textit{neural representation as specification}, as opposed to the traditional \textit{data as specifications}. Specifically, we leverage neural network activation patterns (NAPs) to specify the correct behaviors of neural networks. We argue this could address two major drawbacks of ``data as specifications''. First, NAPs incorporate intrinsic properties of networks which data fails to do. Second, NAPs could cover much larger and more flexible regions compared to $L_\infty$ norm-balls centered around reference points, making them appealing to real-world applications in verifying unseen data. Moreover, we introduce a relaxation factor that specifies the abstraction level of NAPs, which plays an essential role in determining the effectiveness of NAPs as the specification. We also propose a simple method to mine relaxed NAPs and show that working with NAPs can be easily supported by modern neural network verifiers such as Marabou. Through a simple case study and thorough valuation on the MNIST and CIFAR benchmarks, we show that using NAPs as the specification not only addresses major drawbacks of \textit{data as specifications}, but also demonstrates important properties such as non-ambiguity and one order of magnitude stronger verifiable bounds. We foresee that NAPs have the great potential of serving as simple, reliable, and efficient certificates for neural network predictions. 

\newpage
\bibliographystyle{iclr2023_conference}
\bibliography{main}

\newpage

\appendix
% \section{Appendix}
\label{Appendix}

\section{A running example}
% We introduce a simple NN to help with illustrating later concepts.
To help with illustrating later ideas, we present a two-layer feed-forward neural network XNET (\Cref{fig:xornet}) to approximate an analog XOR function $f(x_0, x_1):[[0,0.3] \cup [0.7, 1]]^2 \rightarrow \{0, 1\}$ such that  $f(x_0, x_1) = 1$ iff $(x_0 \leq 0.3 \wedge x_1 \geq 0.7)$ or $(x_0 \geq 0.7 \wedge x_1 \leq 0.3)$. The network computes the function
\begin{align*}
    F_\text{XNET}(x) = \bW^1\max(\bW^0(x)+\bb^0, 0) + \bb^1
\end{align*}
where $x=[x_0, x_1]$, and values of $\bW^0, \bW^1, \bb^0, \bb^1$ are shown in edges of \Cref{fig:xornet}. $\cC(x) = 0$ if $F_\text{XNET}(x)[0]>F_\text{XNET}(x)[1]$, $\cC(x)=1$ otherwise.

Note that the network is not arbitrary. We have obtained it by constructing two sets of 1\,000 randomly generated inputs, and training on one and validating on the other until the NN achieved a perfect F1-score of 1. 
\begin{figure}[t]
     \centering
     \begin{subfigure}[b]{0.2\textwidth}
         \centering
         \includegraphics[width=0.8\textwidth]{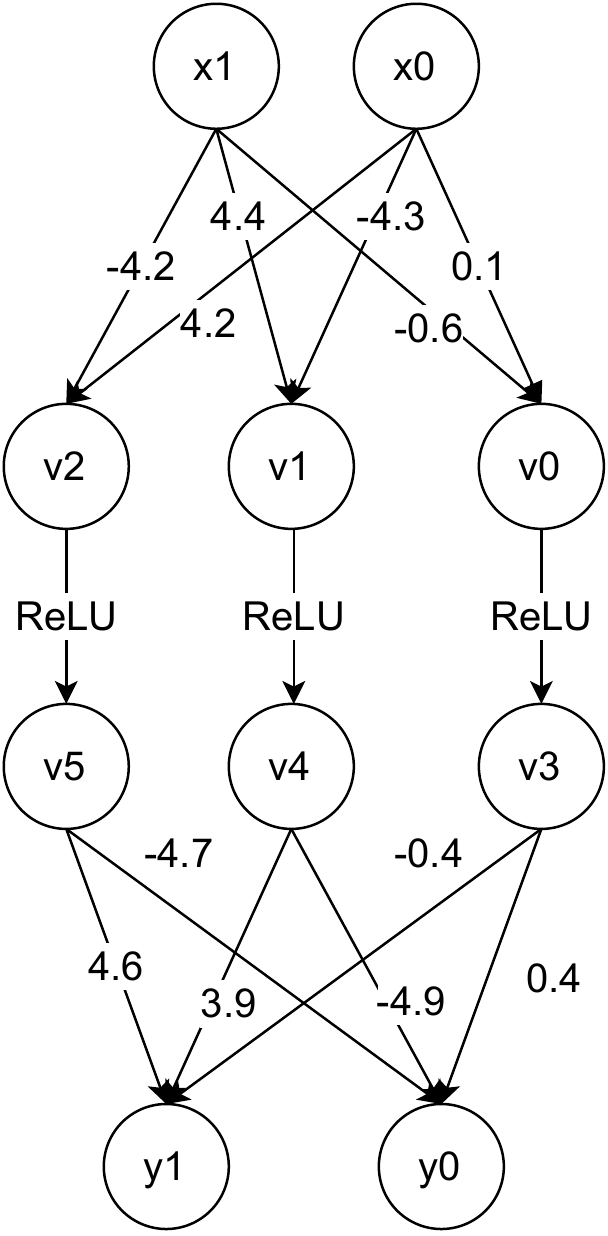}
         \caption{XNET: A NN that computes the analog XOR function.}
         \label{fig:xornet}
     \end{subfigure}
     \hfill
     \begin{subfigure}[b]{0.35\textwidth}
         \centering
         \tiny
         \begin{align*}
             v_0 &= 0.1x_0 - 0.6 x_1 \\
             v_1 &= -4.3x_0 + 4.4 x_1 \\
             v_2 &= 4.2x_0 -4.2x_1 \\
             v_3 &= max(v_0, 0)\\
             v_4 &= max(v_1, 0)\\
             v_5 &= max(v_2, 0)\\
             y_0 &= 0.4v_3 -4.9v_4+3.9v_5+6.7 \\
             y_1 &= -0.4v_3 + 3.9v_4 + 4.6v_5-7.4\\
        % \end{align*}
        % \begin{align*}
             x_0 &\leq 0.1 \wedge x_0 \geq 0.02 \\
             x_1 &\leq 0.1 \wedge x_1 \geq 0.02 \\
             0 &< y_0 - y_1\\
         \end{align*}
         \caption{Marabou's system of constraints for verifying that XNET is 0.04-robust at (0.06, 0.06)}
         \label{fig:marabou_consts}
     \end{subfigure}
     \hfill
     %-------------------------------------
    \begin{subfigure}[b]{0.35\textwidth}
         \centering
         \tiny
         \begin{align*}
             v_0 &= 0.1x_0 - 0.6 x_1 \\
             v_1 &= -4.3x_0 + 4.4 x_1 \\
             v_2 &= 4.2x_0 -4.2x_1 \\
             v_3 &= v_0 \\
             v_4 &= max(v_1, 0) \\
             v_5 &= 0 \\
             y_0 &= 0.4v_4 -4.9v_5+3.9v_6+6.7 \\
             y_1 &= -0.4v_4 + 3.9v_5 + 4.6v_6-7.4 \\
            x_0 &\leq 0.3 \wedge x_0 \geq 0 \\
             x_1 &\leq 0.3 \wedge x_1 \geq 0 \\
             v_0 &\geq 0 \\
             v_2 &\leq 0
         \end{align*}
         \caption{Check if $\cP^1=((z_0),())$ and $\cP^0 = ((),(z_2))$ are non-ambiguous in the first quadrant using Marabou}
         \label{fig:marabou_wiht_relu_consts}
     \end{subfigure}
     \hfill
        \caption{Using Marabou to verify NAP properties of XNET.}
        \label{fig: XOR_example}
\end{figure}

\section{Other Evaluations}
\label{appendix:other_evaluations}

\subsection{$L_1$-norms of distance}
\label{appendix:l1_norm}

\begin{figure}[!htb]
         \centering        \includegraphics[width=0.75\textwidth]{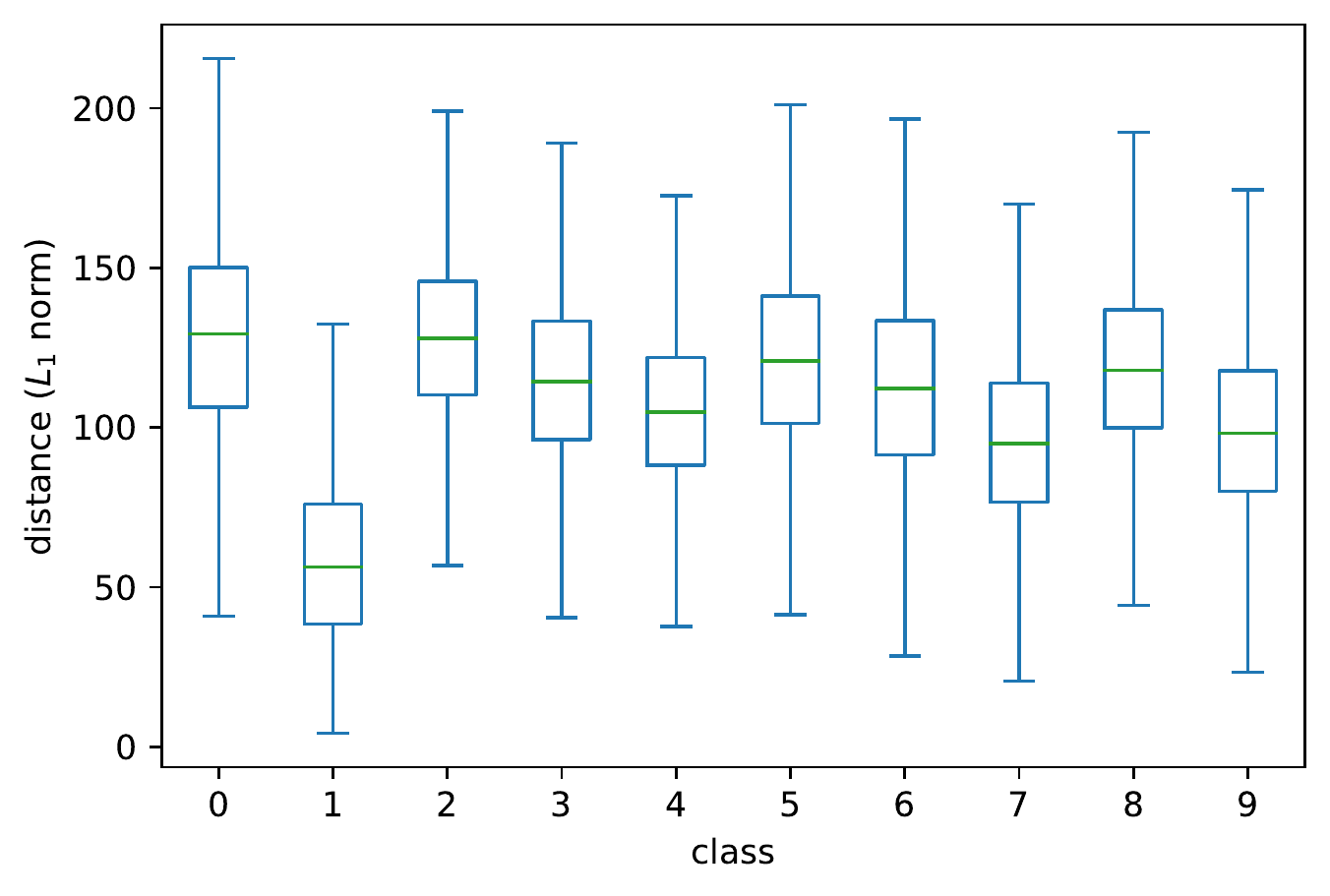}
         \caption{The distribution of $L_1$-norms of all image pairs for each class.}
         \label{fig:mnist_L1}
\end{figure}
\Cref{fig:mnist_L1} shows the distributions of $L_1$-norms of all image pairs from the same class, similar to \cref{fig:mnist_norm}, the distances between image pairs from class 1 are much smaller compare to other classes.

\subsection{Overlap ratio}
\label{appendix:overlap_ratio}

\begin{table}[th!]
\caption{The maximum overlap ratio for each label (class) on a given $\delta.\NAP$ for MNIST. Each cell is obtained by $\max_i |N_{col}^{\delta} \bigcap N_i^{\delta}|/ |N_{col}^{\delta}|$ where $N_{col}^{\delta}$ is the set of neurons in the dominant pattern for the label (class) in the header of the column of the selected cell with the given $\delta$, $N_i$ is the set of neurons in the dominant pattern for the label (class) $i$ with the given $\delta$.} 
\label{tab:mnist_max_overlap_ratio}
\begin{center}
\begin{tabular}{ |l || r r r r r r r r r r| }
\hline
& 0& 1& 2& 3& 4& 5& 6& 7& 8& 9\\
\hline\hline
0.00&0.959 &0.928 &0.963 &0.966 &0.972 &0.973 &0.930 &0.965 &0.957 &0.981\\ 
 \hline
0.01&0.844 &0.834 &0.911 &0.901 &0.881 &0.898 &0.895 &0.884 &0.880 &0.908\\ 
 \hline
0.05&0.864 &0.885 &0.909 &0.904 &0.915 &0.908 &0.899 &0.897 &0.890 &0.893\\ 
 \hline
0.10&0.877 &0.900 &0.910 &0.901 &0.921 &0.910 &0.890 &0.899 &0.900 &0.901\\ 
 \hline
0.15&0.876 &0.904 &0.904 &0.900 &0.919 &0.913 &0.893 &0.907 &0.904 &0.900\\ 
 \hline
0.25&0.893 &0.922 &0.913 &0.912 &0.928 &0.925 &0.905 &0.916 &0.916 &0.913\\ 
 \hline
0.50&0.903 &0.905 &0.925 &0.923 &0.926 &0.923 &0.907 &0.918 &0.927 &0.927\\ 
 \hline
\end{tabular}
\end{center}
\end{table}
\Cref{fig:overlap ratio} shows the heatmap of the overlap ratio between any two classes for 6 $\delta$ values. For the grid in each column in a heatmap, the overlap ratio is calculated by the number of overlapping neurons of the {\NAP}s of the class labelled for the row and the column divided by the number of neurons in the \NAP of the class labelled for the column, which is why the values in the heatmaps are not symmetric along the diagonal. Based on the shade of the colors in our heapmap, we can see that, during the process of decreasing $\delta$, the overlapping ratios decrease first and then increase in general, it might because that, with the loose of restriction on when a neuron is considered as activated/inactivated, more neurons are included in the \NAP, which means more constrains, but at the same time, for two {\NAP}s of any two classes, it is more likely that they have more neurons appearing in both {\NAP}s.

\begin{figure}[!htb]
     \centering
     \begin{subfigure}[t]{0.3\textwidth}
         \centering
         \includegraphics[width=\textwidth]{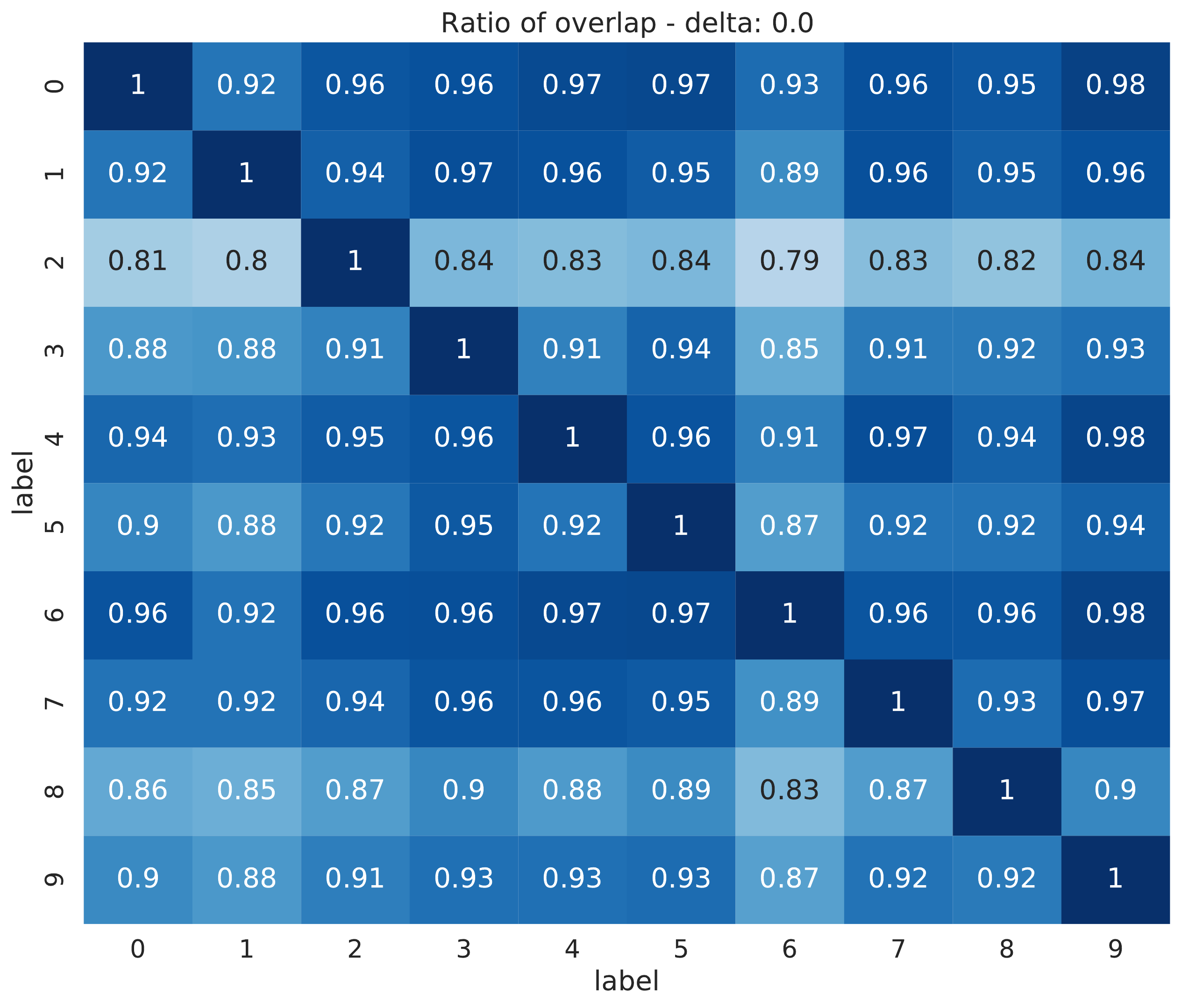}
         \caption{$\delta=0.00$}
         \label{fig:overlap ratio 0}
     \end{subfigure}
     \hfill
     \begin{subfigure}[t]{0.3\textwidth}
         \centering
         \includegraphics[width=\textwidth]{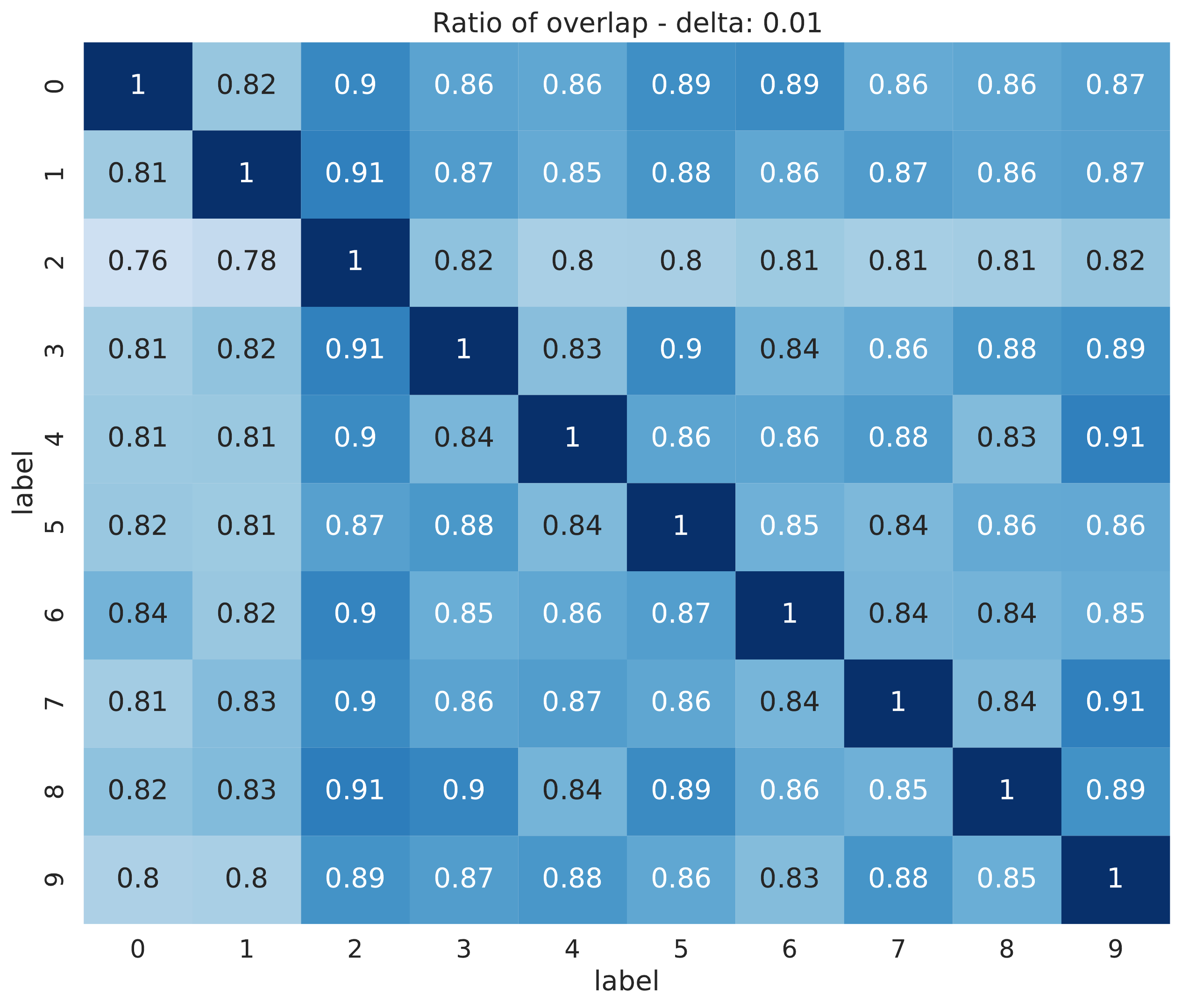}
         \caption{$\delta=0.01$}
         \label{fig:overlap ratio 0.01}
     \end{subfigure}
     \hfill
     \begin{subfigure}[t]{0.345\textwidth}
         \centering
         \includegraphics[width=\textwidth]{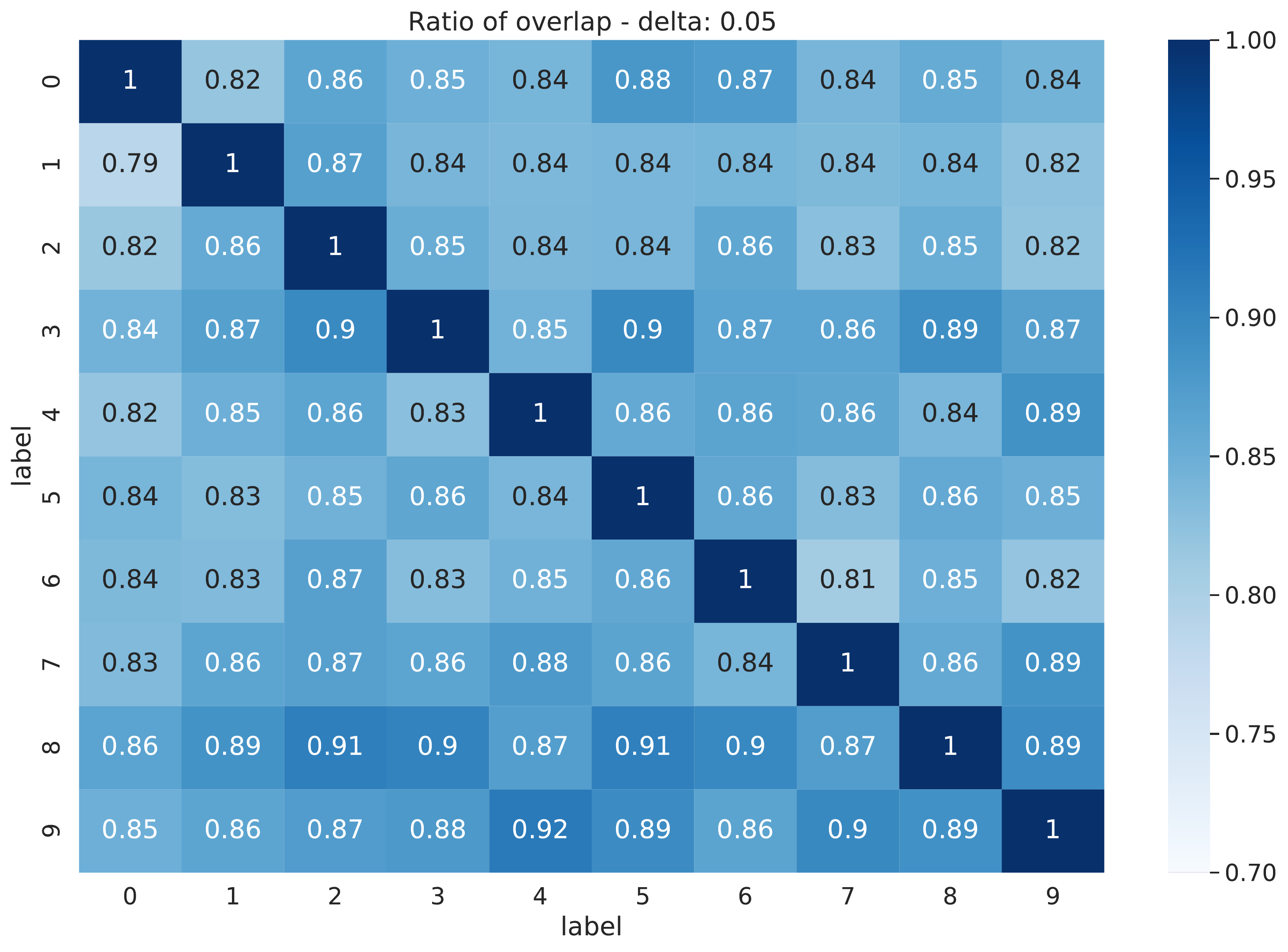}
         \caption{$\delta=0.05$}
         \label{fig:overlap ratio 0.05}
     \end{subfigure}
     \hfill
     \begin{subfigure}[t]{0.3\textwidth}
         \centering
         \includegraphics[width=\textwidth]{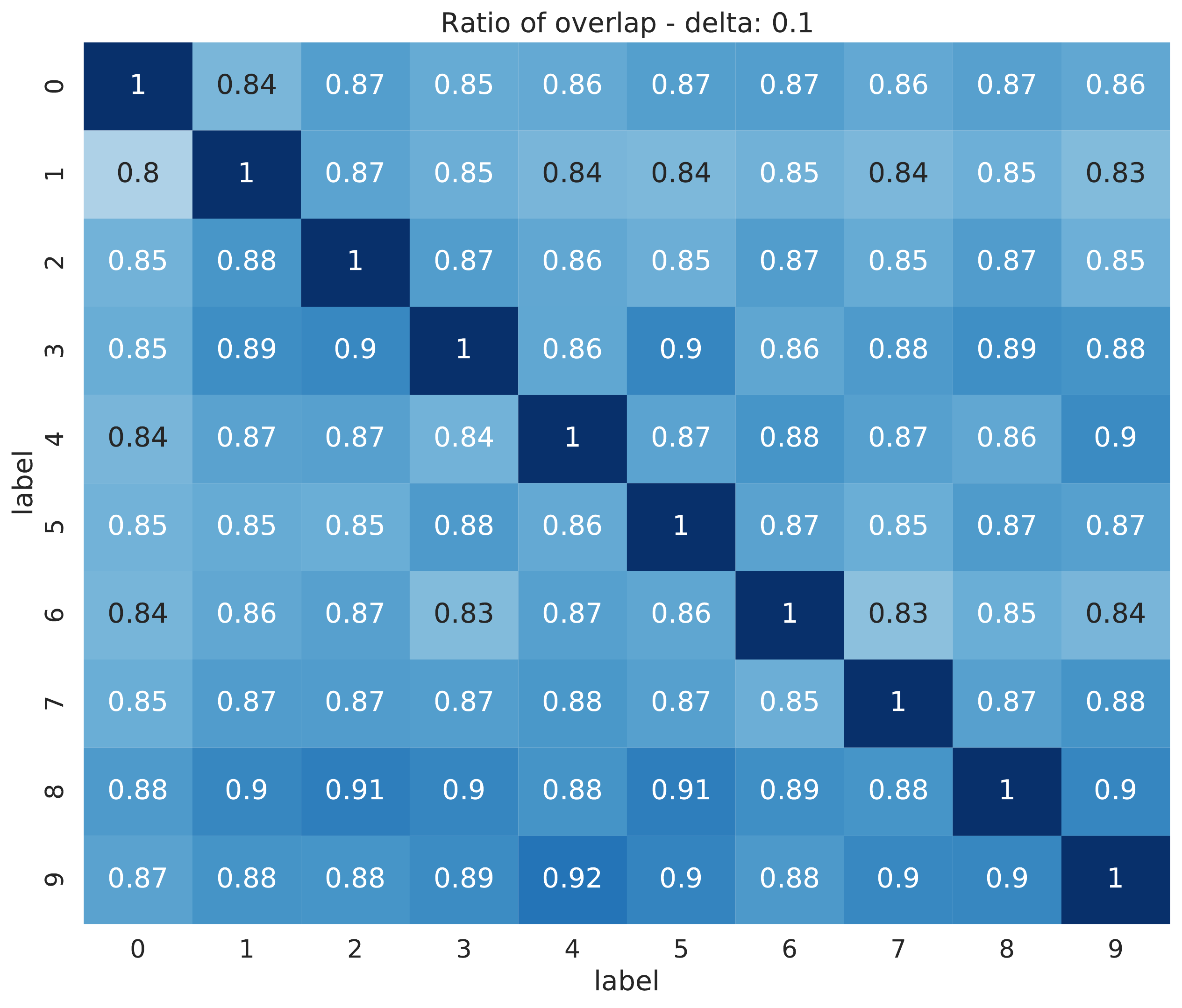}
         \caption{$\delta=0.10$}
         \label{fig:overlap ratio 0.1}
     \end{subfigure}
     \hfill
     \begin{subfigure}[t]{0.3\textwidth}
         \centering
         \includegraphics[width=\textwidth]{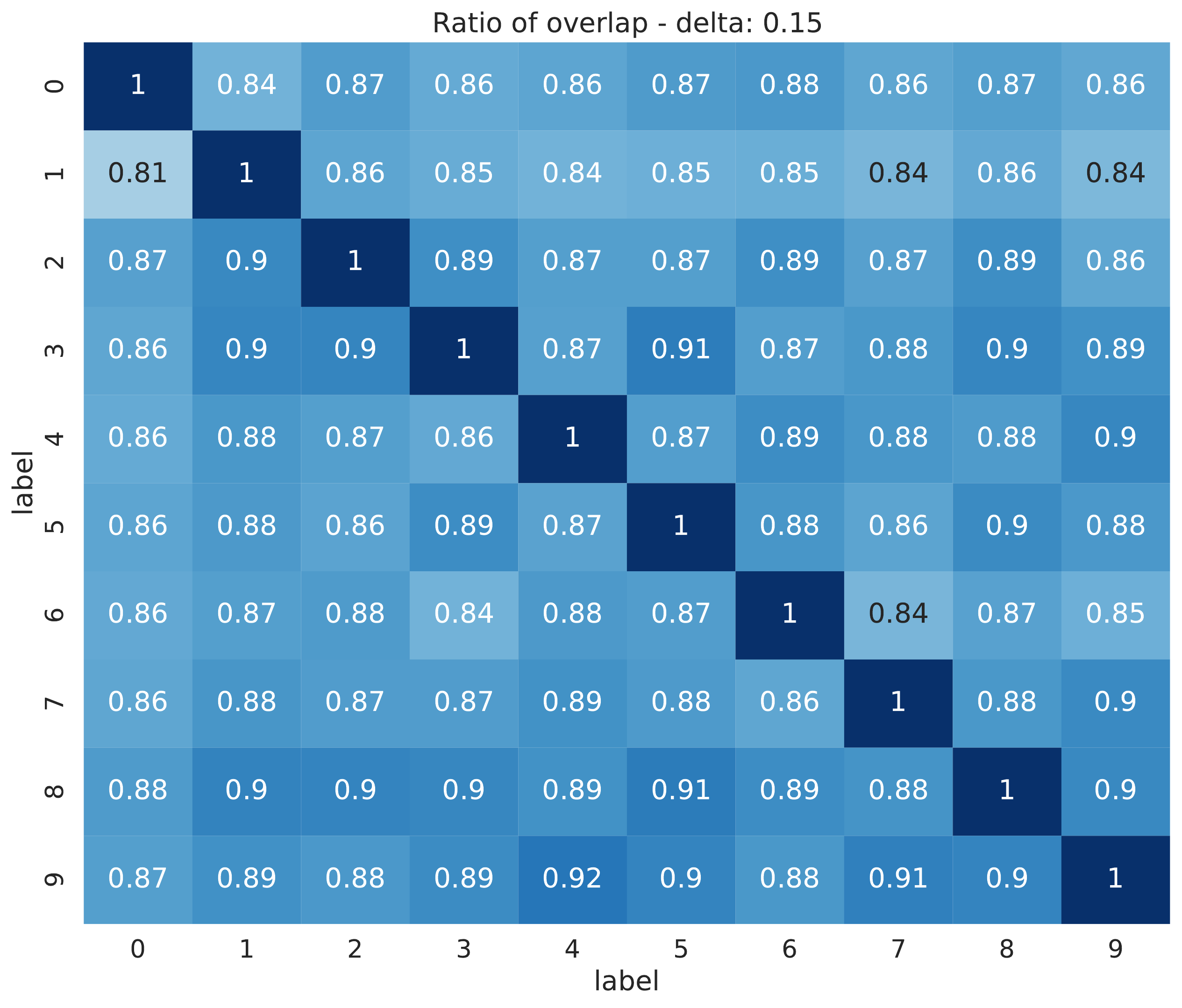}
         \caption{$\delta=0.15$}
         \label{fig:overlap ratio 0.1}
     \end{subfigure}
     \hfill
     \begin{subfigure}[t]{0.345\textwidth}
         \centering
         \includegraphics[width=\textwidth]{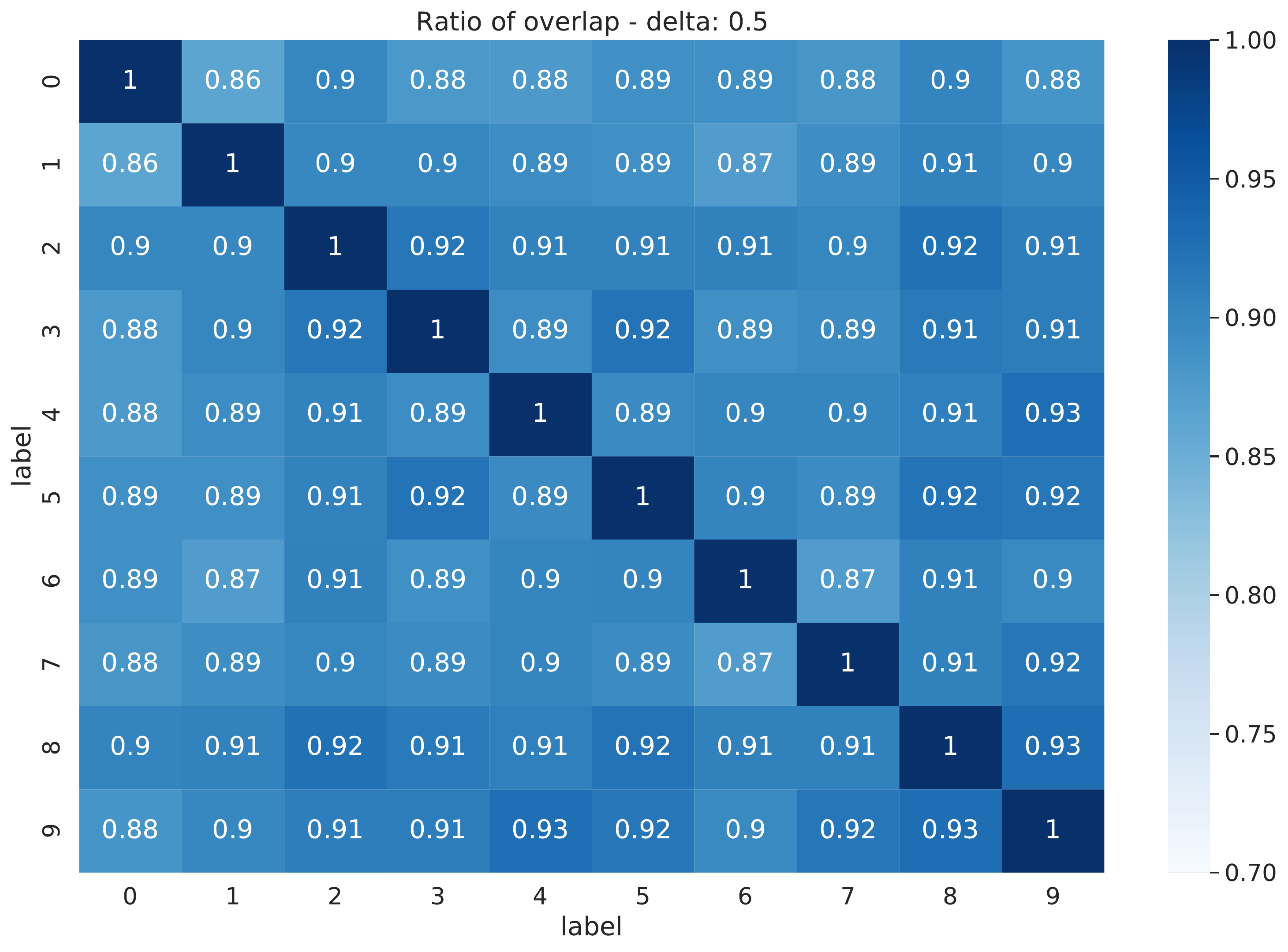}
         \caption{$\delta=0.50$}
         \label{fig:overlap ratio 0.5}
     \end{subfigure}
        \caption{Overlap ratio of the dominant pattern of two labels (classes) on a given $\delta.\NAP$. Values in each grid are obtained by $ |N_{col}^{\delta} \bigcap N_{row}^{\delta}|/ |N_{col}^{\delta}|$ where $N_{col}^{\delta}$ is the set of neurons in the dominant pattern for the label (class) of the column of the selected grid with the given $\delta$, $N_{row}$ is the set of neurons in the dominant pattern for the label (class) of the row of the selected grid with the given $\delta$.  }
        \label{fig:overlap ratio}
\end{figure}
\cref{tab:mnist_max_overlap_ratio} shows the maximum overlap ratio for one class, that is, for one reference class, the maximum overlap ratio between this reference class and any other class. This table is basically extracting the maximum values of each column other than the 1 on the diagonal in our heatmap in \cref{fig:overlap ratio}, in the each column of our table, it also follows the pattern that the value of overlap ratio decreases first and then increase with the decrease of $\delta$.

\section{Misclassification Examples}
\label{appendix:misclassification}

In this section, we display some interesting exmaples from the the MNIST test set that follow the \NAP of some class other than their ground truth, which means these images are misclassified. We consider these samples interesting because, instead of misclassification, it is more reasonable to say that these images are given wrong ground truth from human perspective.
% \cref{fig:Misclassify_to_1} shows some interesting samples from the MNIST test set that follow the \NAP of some class other than their ground truth, i.e. these images are misclassified. We consider these samples interesting because instead of misclassification, from human perspective, it is more reasonable to say that these images are given wrong ground truth, especially for \cref{fig:misclsfy_6_to_1} and \cref{fig:misclsfy_7_to_1} which is very likely to be recognized as 1 instead of 6 and 7 respectively from human view. As for \cref{fig:misclsfy_9_to_1}, humans might still be able to recognize it as 9, however, it is very understandable that this 9 could easily trick the model to classify it as 1 based on its shape.

\begin{figure}[th]
     \centering
     \begin{subfigure}[t]{0.3\textwidth}
         \centering
         \includegraphics[width=\textwidth]{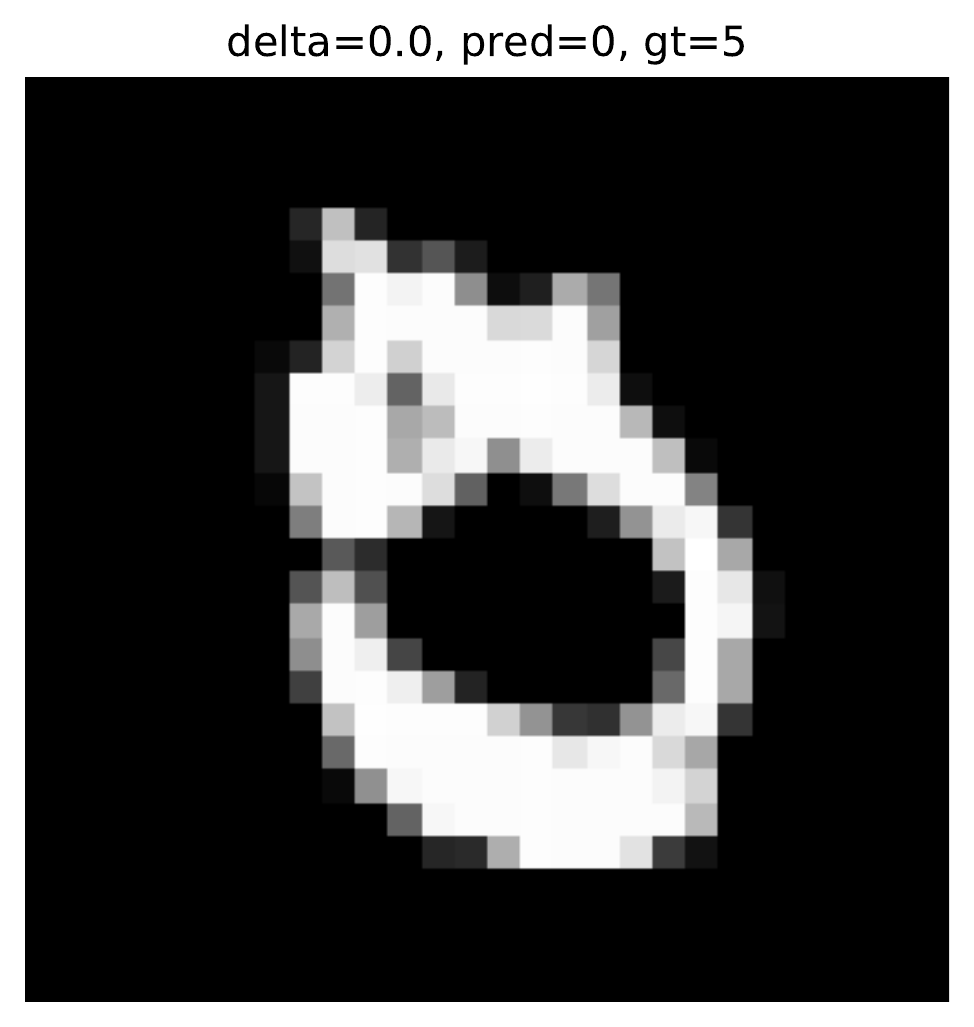}
         \caption{A testing image from MNIST with ground truth 5, classified as 0}
         \label{fig:misclsfy_5_to_0}
     \end{subfigure}
     \hfill
     \begin{subfigure}[t]{0.3\textwidth}
         \centering
         \includegraphics[width=\textwidth]{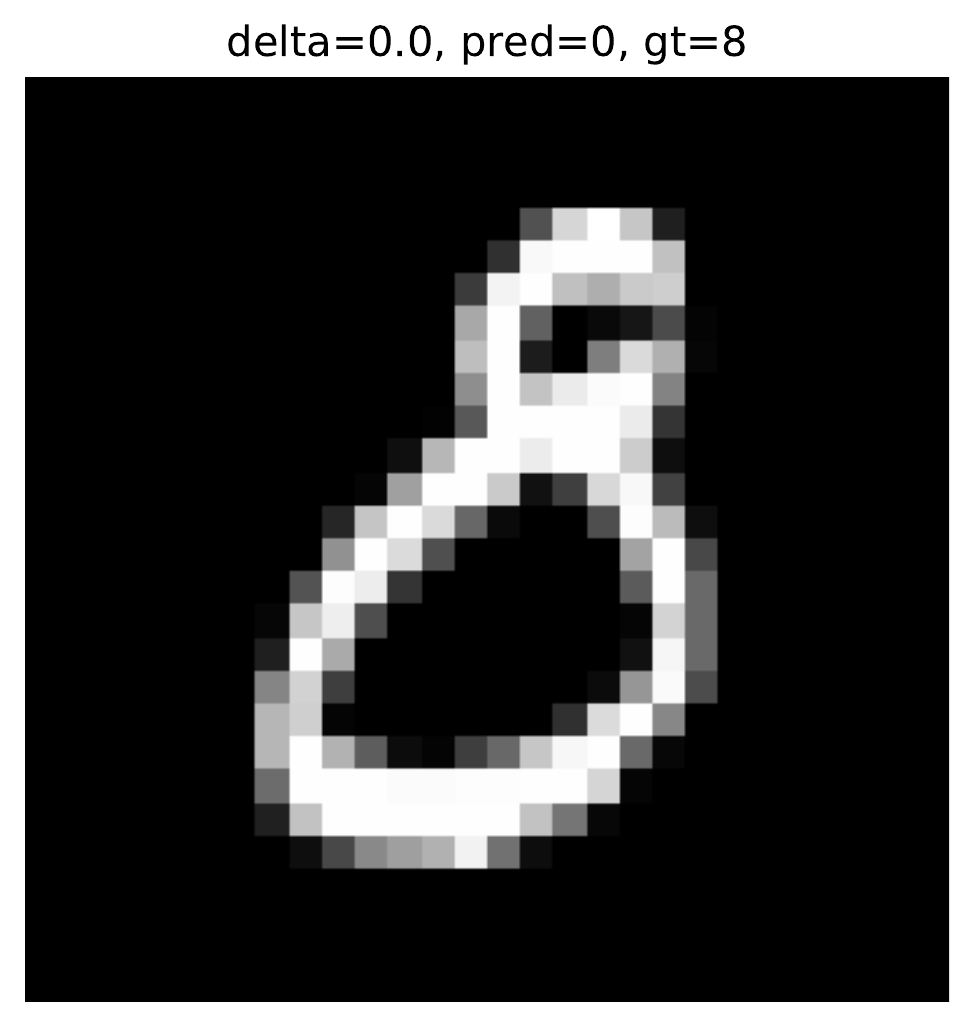}
         \caption{A testing image from MNIST with ground truth 8, classified as 0}
         \label{fig:misclsfy_8_to_0}
     \end{subfigure}
     \hfill
     \begin{subfigure}[t]{0.3\textwidth}
         \centering
         \includegraphics[width=\textwidth]{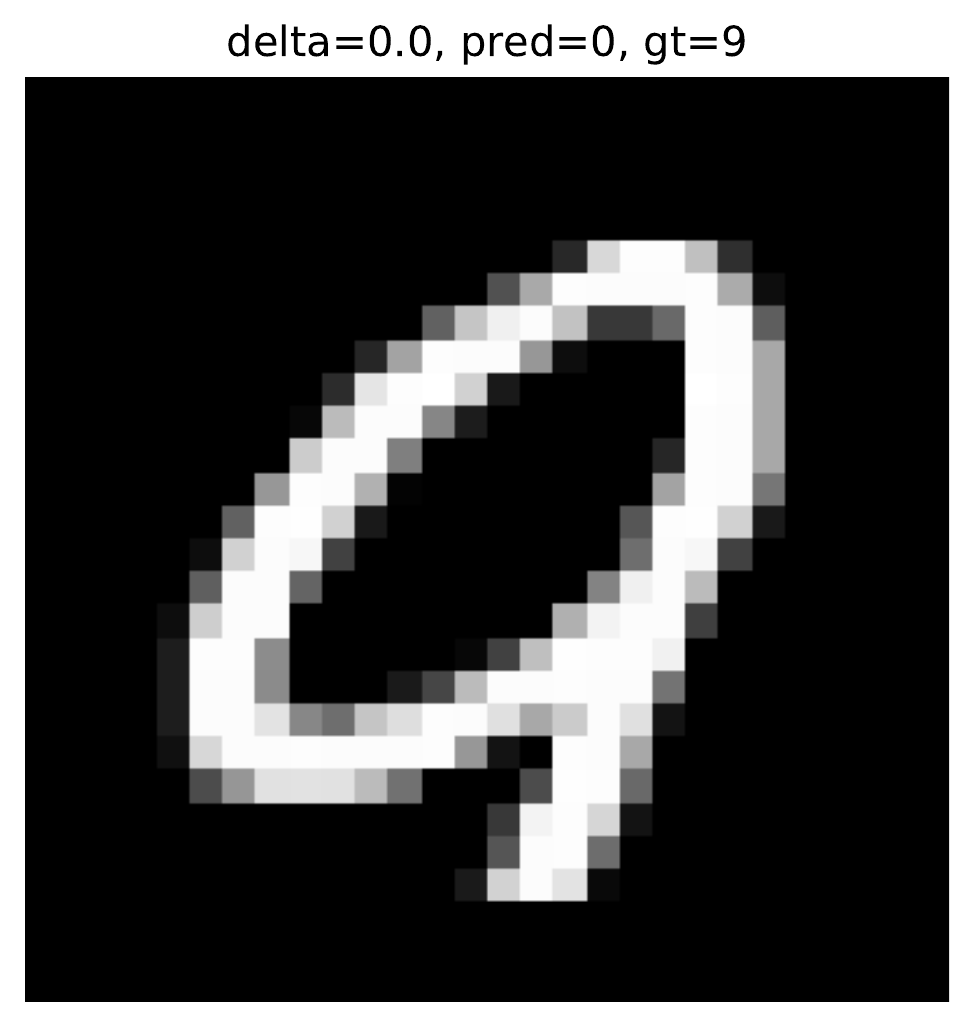}
         \caption{A testing image from MNIST with ground truth 9, classified as 0}
         \label{fig:misclsfy_9_to_0}
     \end{subfigure}
        \caption{Some interesting test images from MNIST that are misclassified as 0 and also follow the $\NAP$ of class 0.}
        \label{fig:Misclassify_to_0}
\end{figure}
\begin{figure}[th]
     \centering
     \begin{subfigure}[t]{0.3\textwidth}
         \centering
         \includegraphics[width=\textwidth]{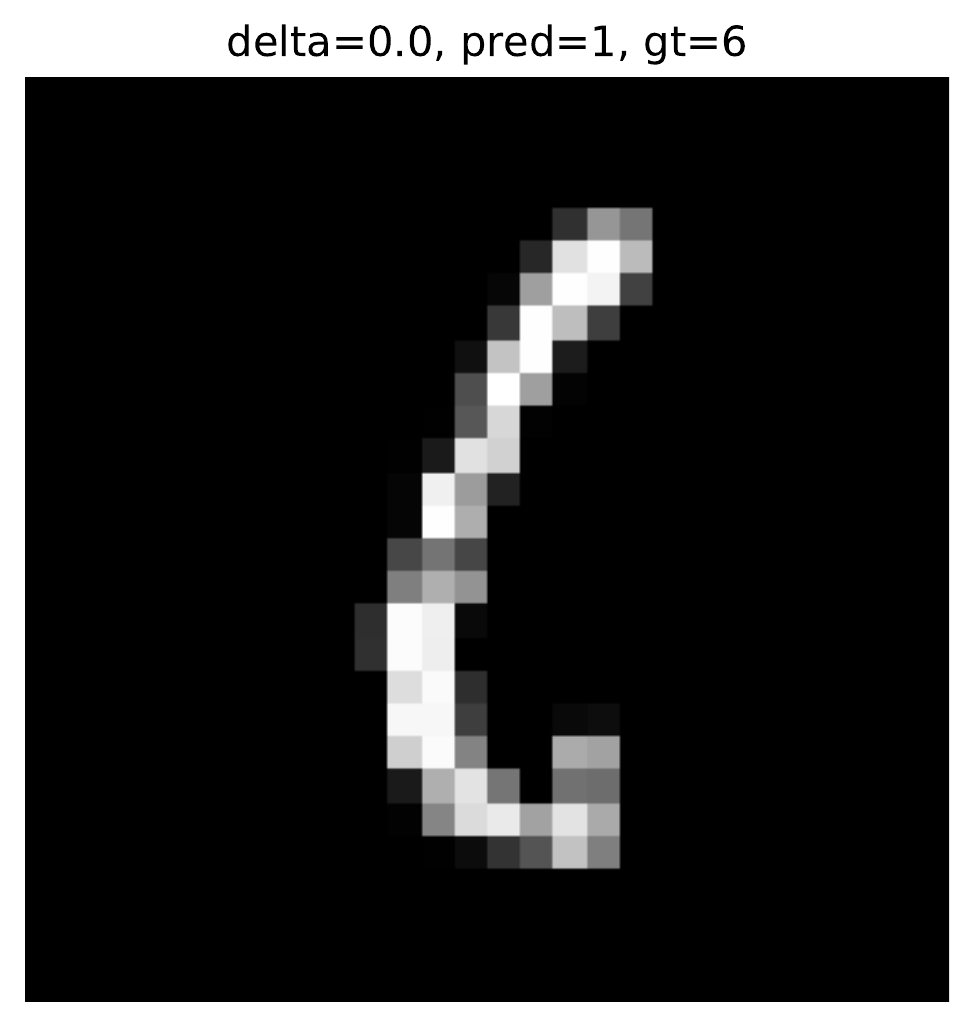}
         \caption{A testing image from MNIST with ground truth 6, classified as 1}
         \label{fig:misclsfy_6_to_1}
     \end{subfigure}
     \hfill
     \begin{subfigure}[t]{0.3\textwidth}
         \centering
         \includegraphics[width=\textwidth]{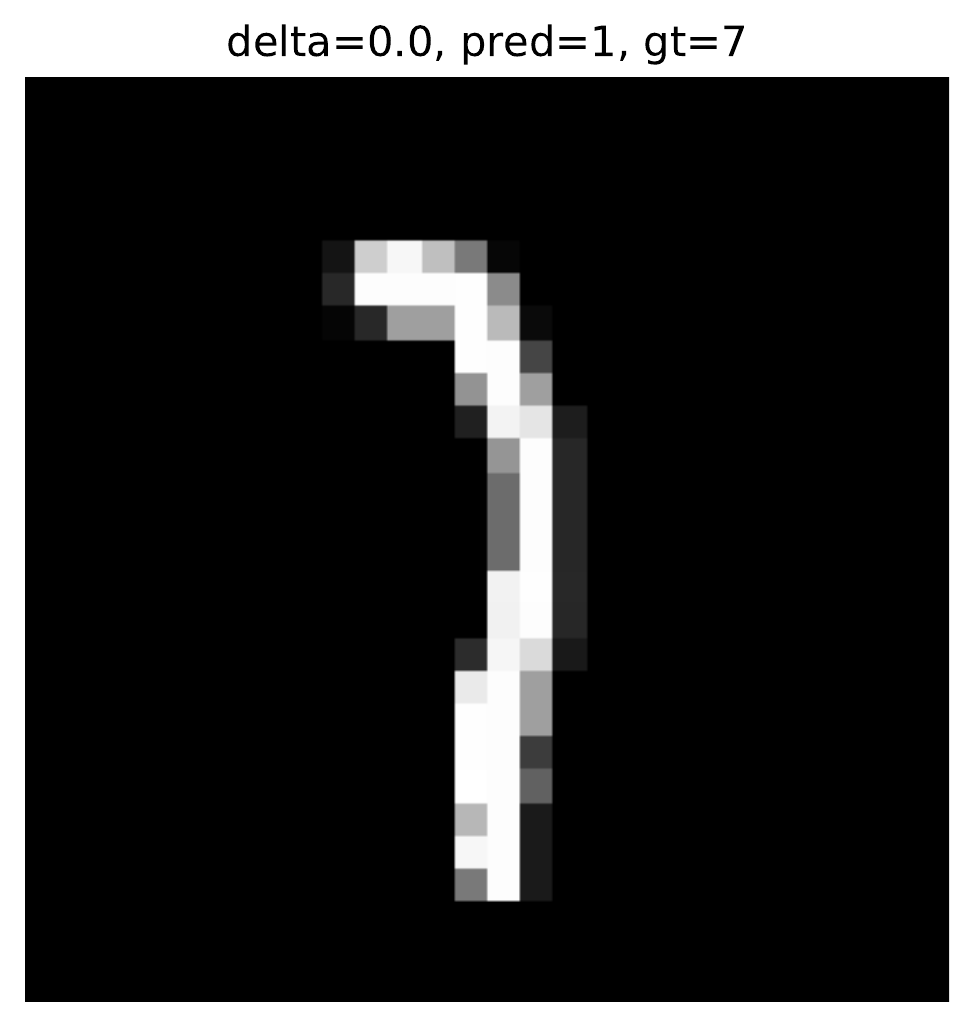}
         \caption{A testing image from MNIST with ground truth 7, classified as 1}
         \label{fig:misclsfy_7_to_1}
     \end{subfigure}
     \hfill
     \begin{subfigure}[t]{0.3\textwidth}
         \centering
         \includegraphics[width=\textwidth]{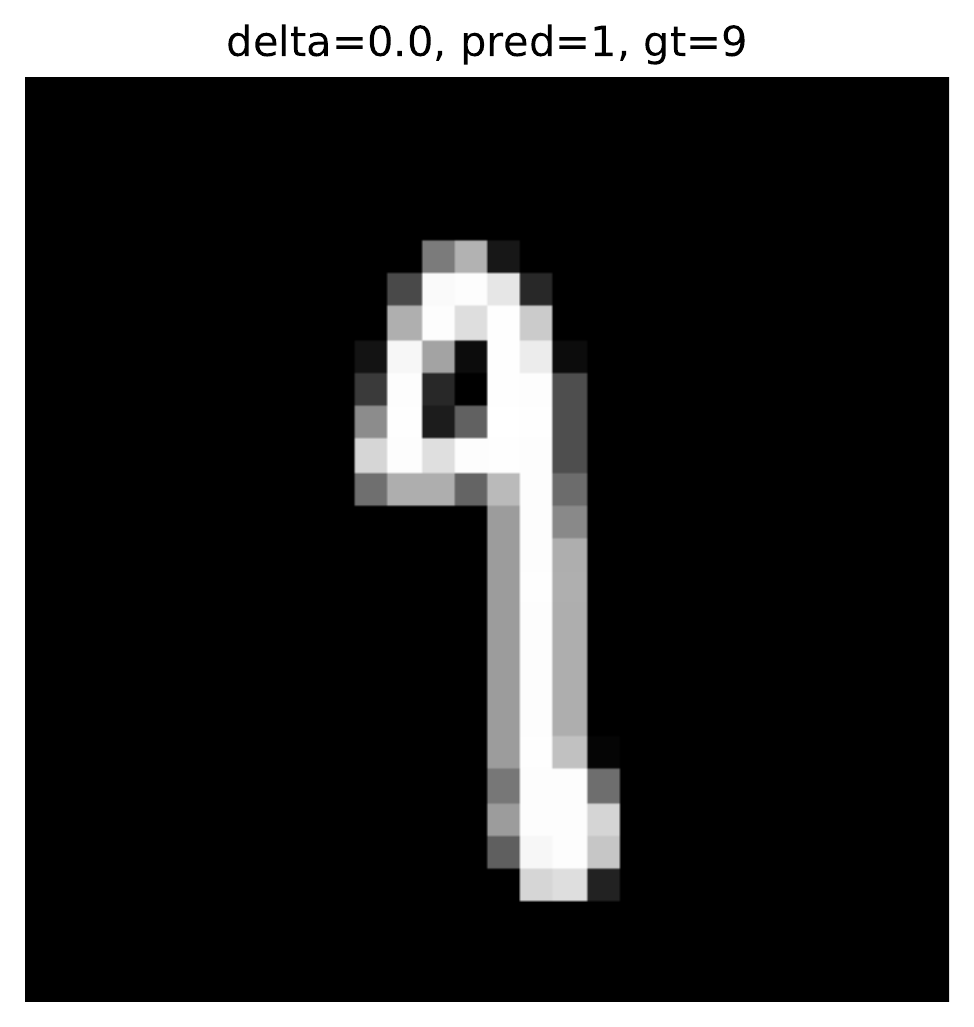}
         \caption{A testing image from MNIST with ground truth 9, classified as 1}
         \label{fig:misclsfy_9_to_1}
     \end{subfigure}
        \caption{Some interesting test images from MNIST that are misclassified as 1 and also follow the $\NAP$ of class 1.}
        \label{fig:Misclassify_to_1}
\end{figure}
\begin{figure}[th]
     \centering
     \begin{subfigure}[t]{0.3\textwidth}
         \centering
         \includegraphics[width=\textwidth]{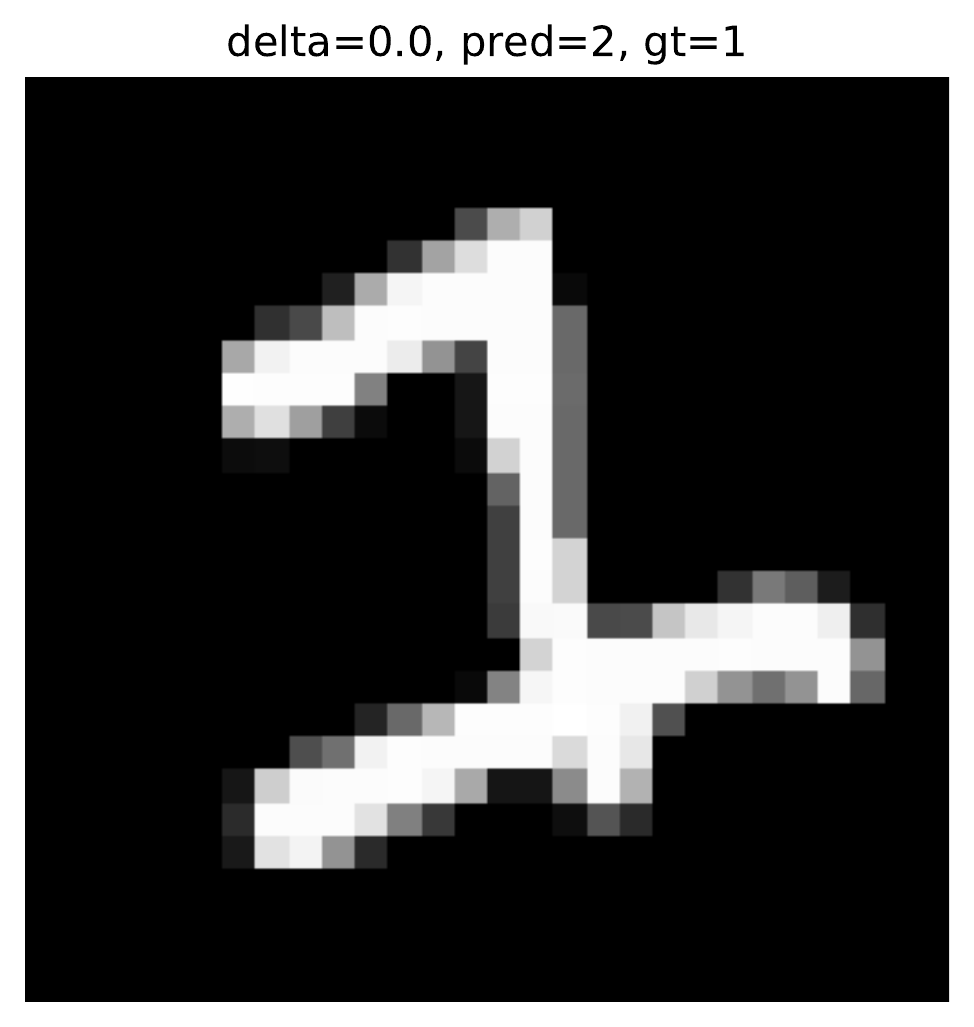}
         \caption{A testing image from MNIST with ground truth 1, classified as 2}
         \label{fig:misclsfy_1_to_2}
     \end{subfigure}
     \hfill
     \begin{subfigure}[t]{0.3\textwidth}
         \centering
         \includegraphics[width=\textwidth]{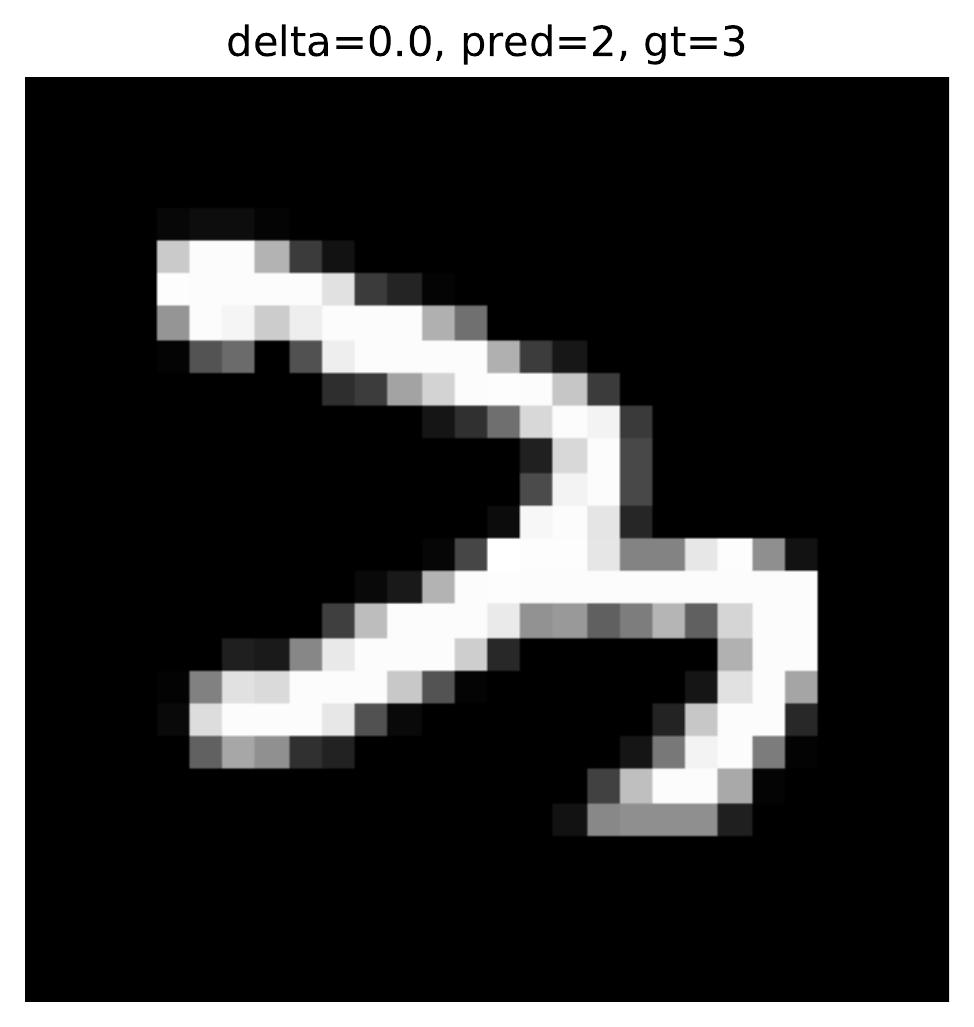}
         \caption{A testing image from MNIST with ground truth 3, classified as 2}
         \label{fig:misclsfy_3_to_2}
     \end{subfigure}
     \hfill
     \begin{subfigure}[t]{0.3\textwidth}
         \centering
         \includegraphics[width=\textwidth]{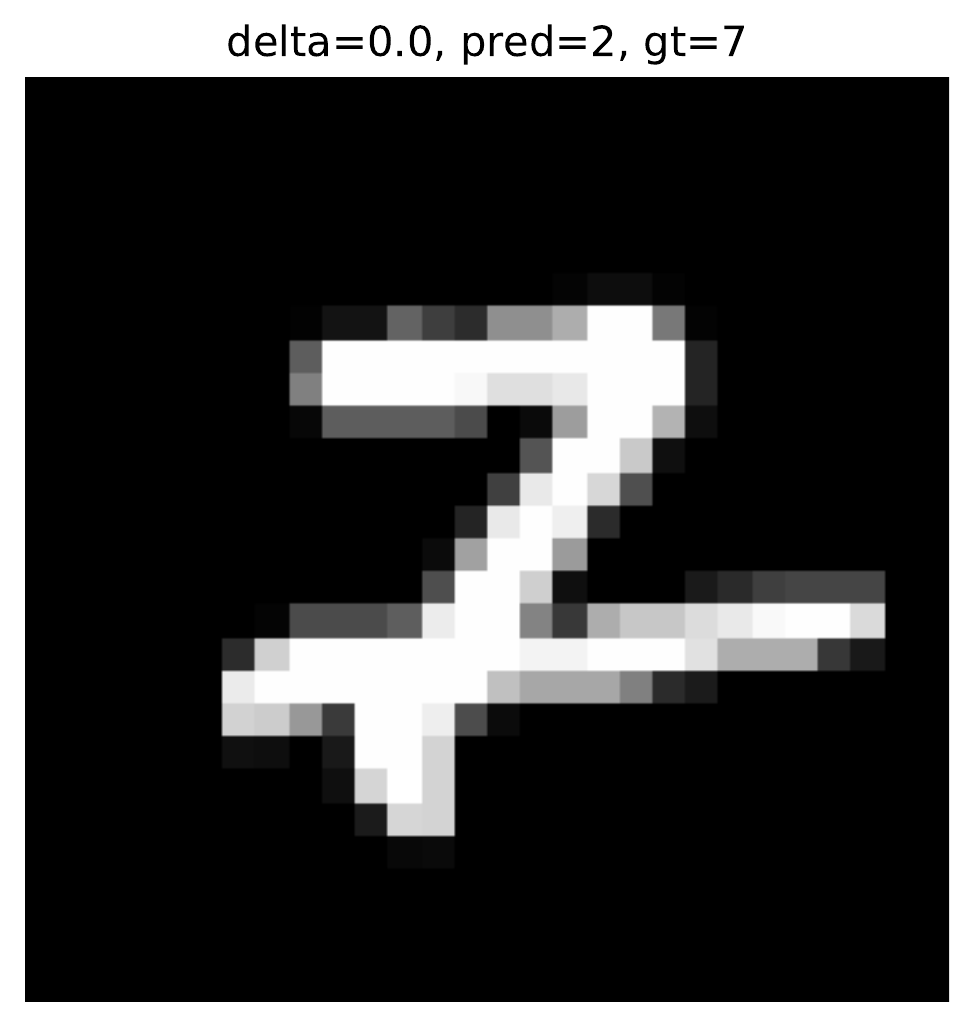}
         \caption{A testing image from MNIST with ground truth 7, classified as 2}
         \label{fig:misclsfy_7_to_2}
     \end{subfigure}
        \caption{Some interesting test images from MNIST that are misclassified as 2 and also follow the $\NAP$ of class 2.}
        \label{fig:Misclassify_to_2}
\end{figure}
\begin{figure}[th]
     \centering
     \begin{subfigure}[t]{0.3\textwidth}
         \centering
         \includegraphics[width=\textwidth]{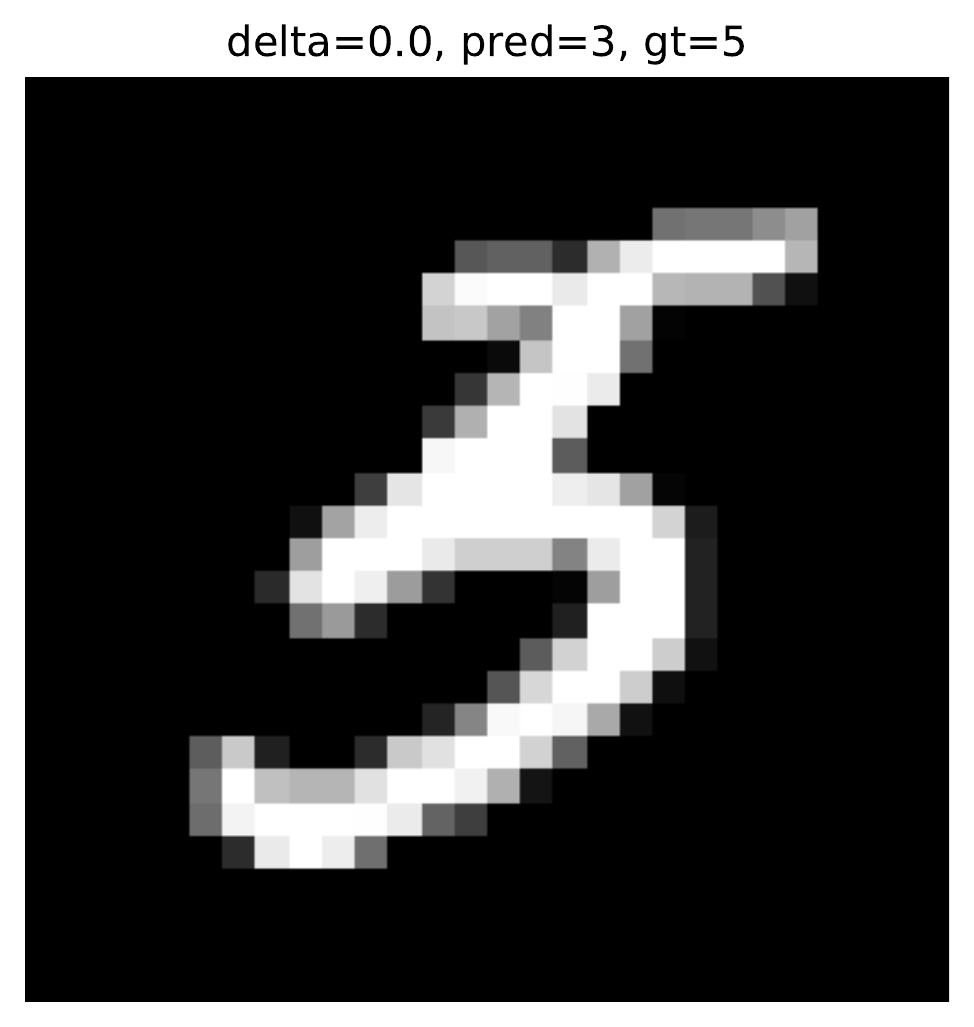}
         \caption{A testing image from MNIST with ground truth 5, classified as 3}
         \label{fig:misclsfy_5_to_3}
     \end{subfigure}
     \hfill
     \begin{subfigure}[t]{0.3\textwidth}
         \centering
         \includegraphics[width=\textwidth]{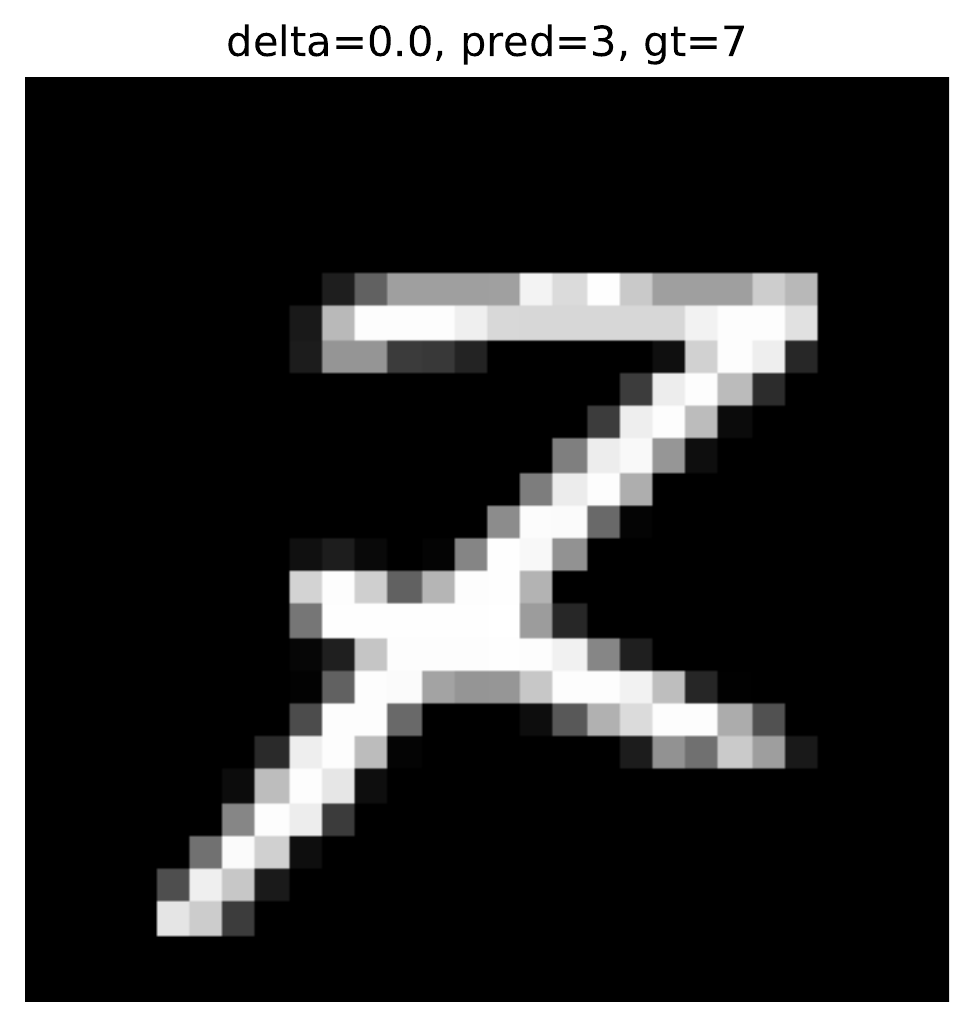}
         \caption{A testing image from MNIST with ground truth 7, classified as 3}
         \label{fig:misclsfy_7_to_3}
     \end{subfigure}
     \hfill
     \begin{subfigure}[t]{0.3\textwidth}
         \centering
         \includegraphics[width=\textwidth]{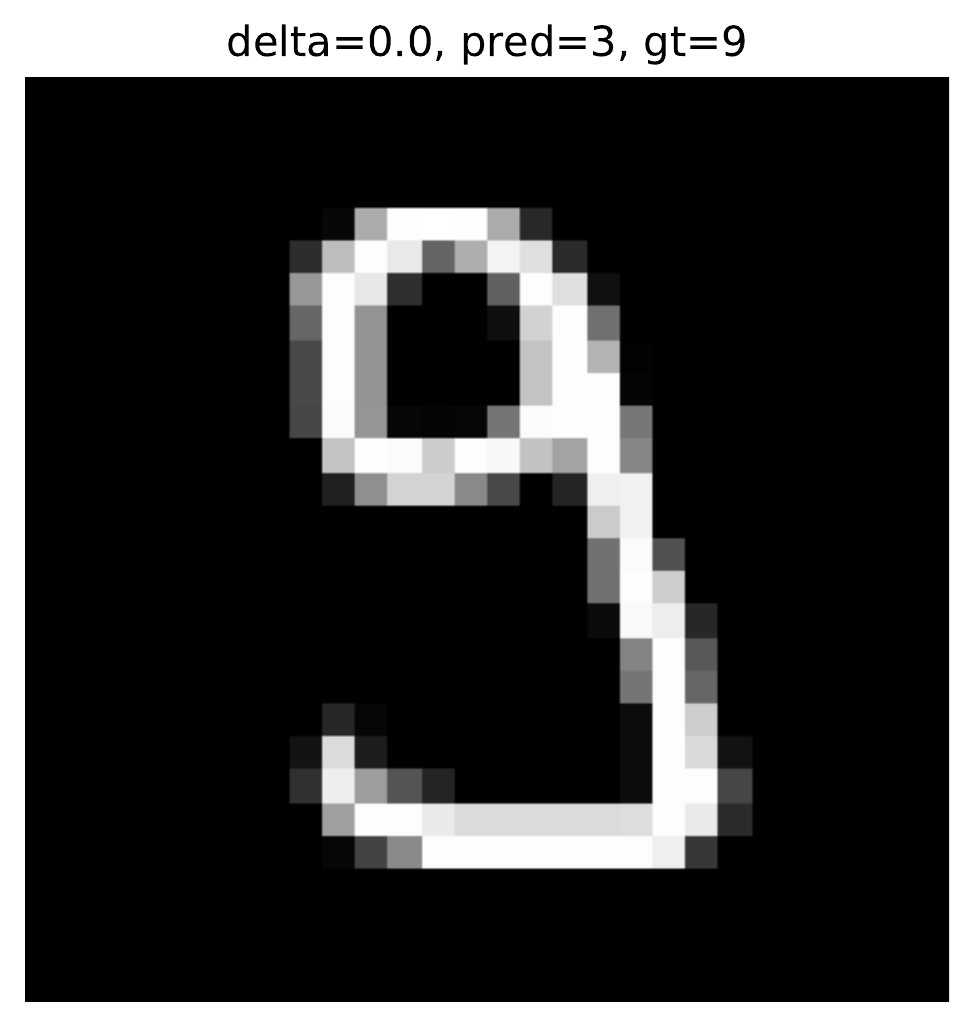}
         \caption{A testing image from MNIST with ground truth 9, classified as 3}
         \label{fig:misclsfy_9_to_3}
     \end{subfigure}
        \caption{Some interesting test images from MNIST that are misclassified as 3 and also follow the $\NAP$ of class 3.}
        \label{fig:Misclassify_to_3}
\end{figure}
\begin{figure}[th]
     \centering
     \begin{subfigure}[t]{0.3\textwidth}
         \centering
         \includegraphics[width=\textwidth]{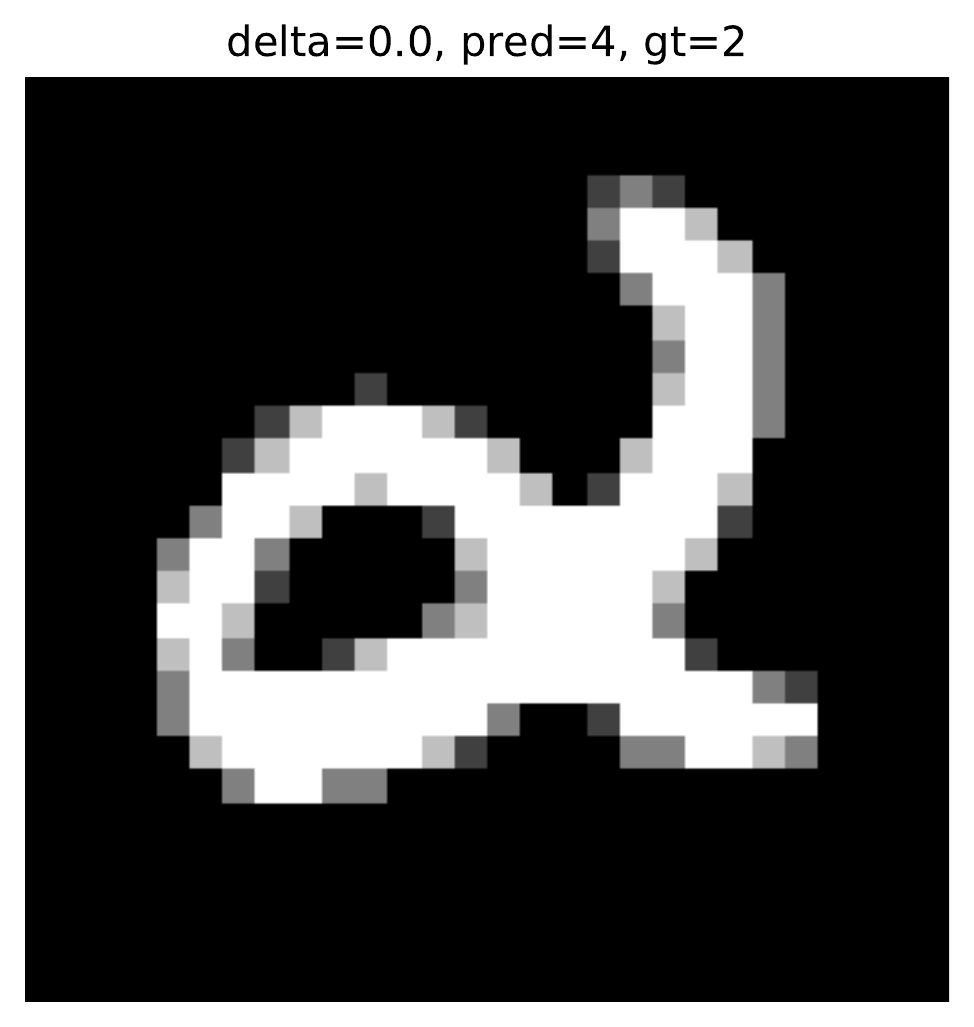}
         \caption{A testing image from MNIST with ground truth 2, classified as 4}
         \label{fig:misclsfy_2_to_4}
     \end{subfigure}
     \hfill
     \begin{subfigure}[t]{0.3\textwidth}
         \centering
         \includegraphics[width=\textwidth]{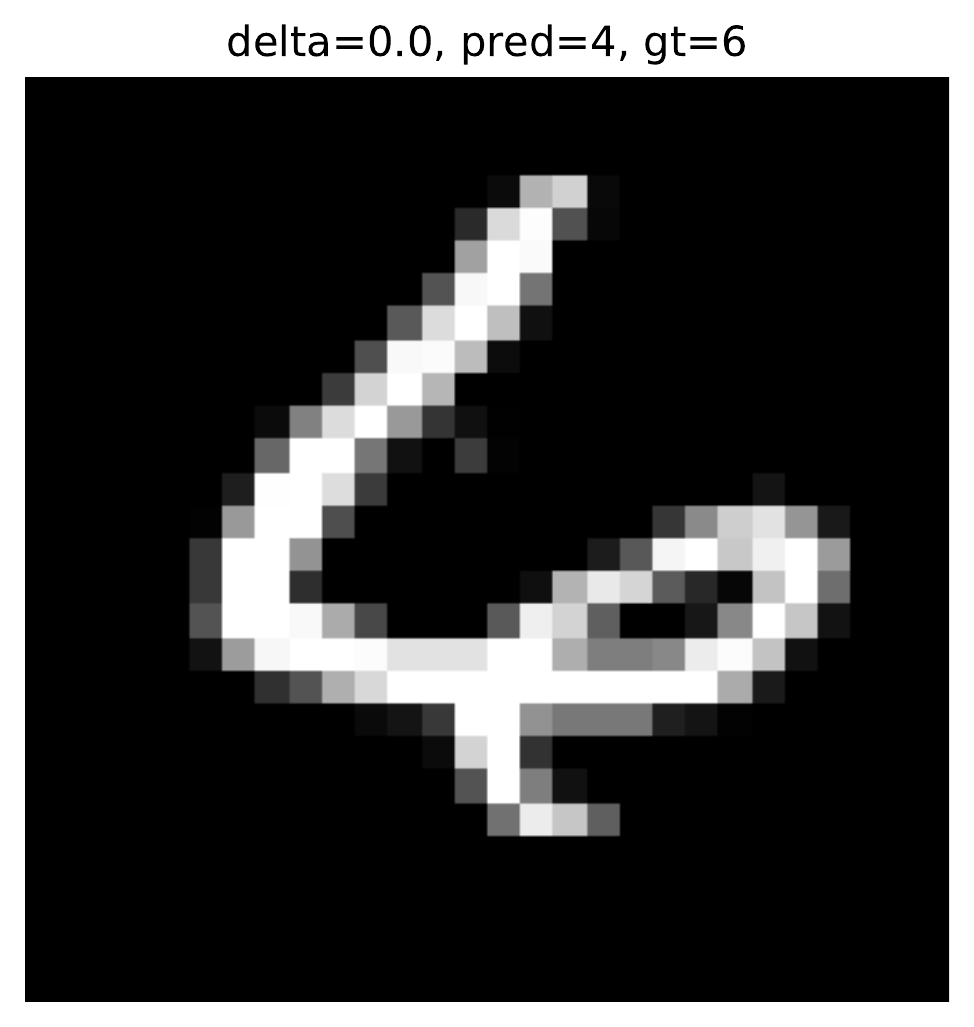}
         \caption{A testing image from MNIST with ground truth 6, classified as 4}
         \label{fig:misclsfy_6_to_4}
     \end{subfigure}
     \hfill
     \begin{subfigure}[t]{0.3\textwidth}
         \centering
         \includegraphics[width=\textwidth]{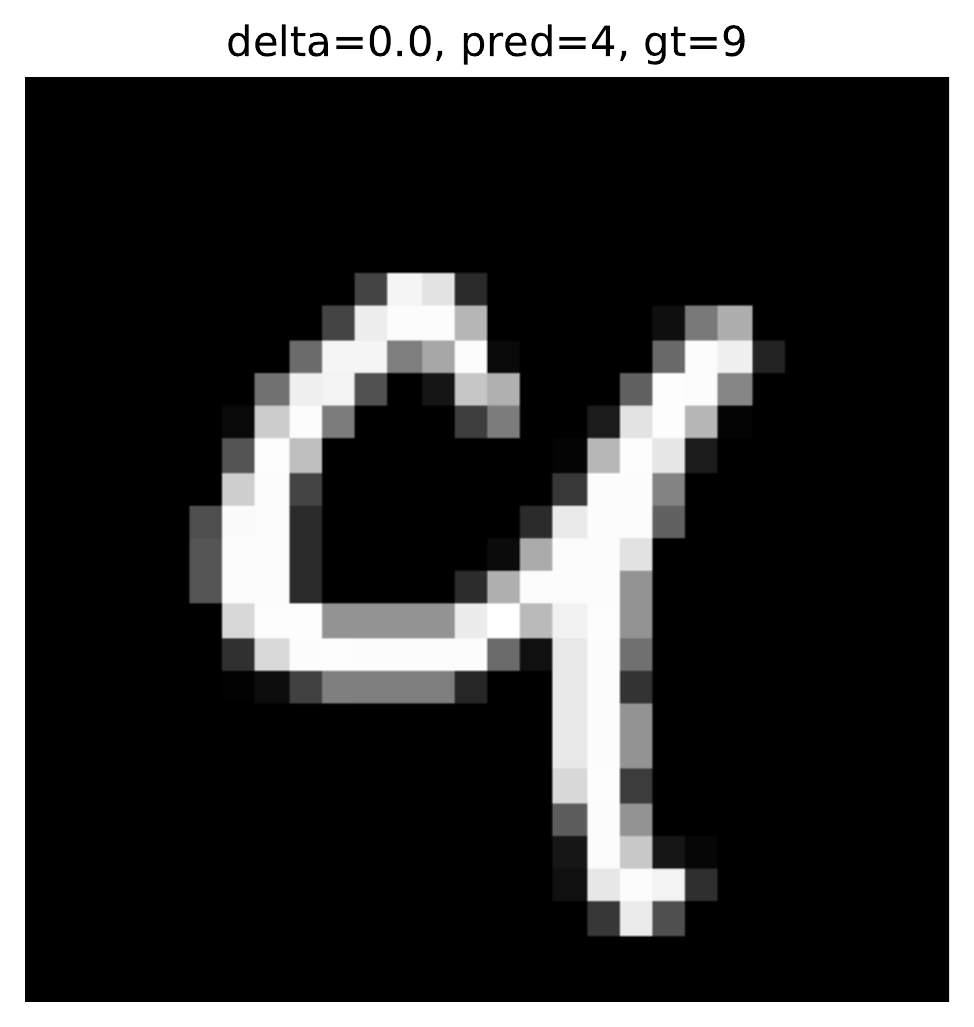}
         \caption{A testing image from MNIST with ground truth 9, classified as 4}
         \label{fig:misclsfy_7_to_2}
     \end{subfigure}
        \caption{Some interesting test images from MNIST that are misclassified as 4 and also follow the $\NAP$ of class 4.}
        \label{fig:Misclassify_to_4}
\end{figure}
\begin{figure}[th]
     \centering
     \begin{subfigure}[t]{0.3\textwidth}
         \centering
         \includegraphics[width=\textwidth]{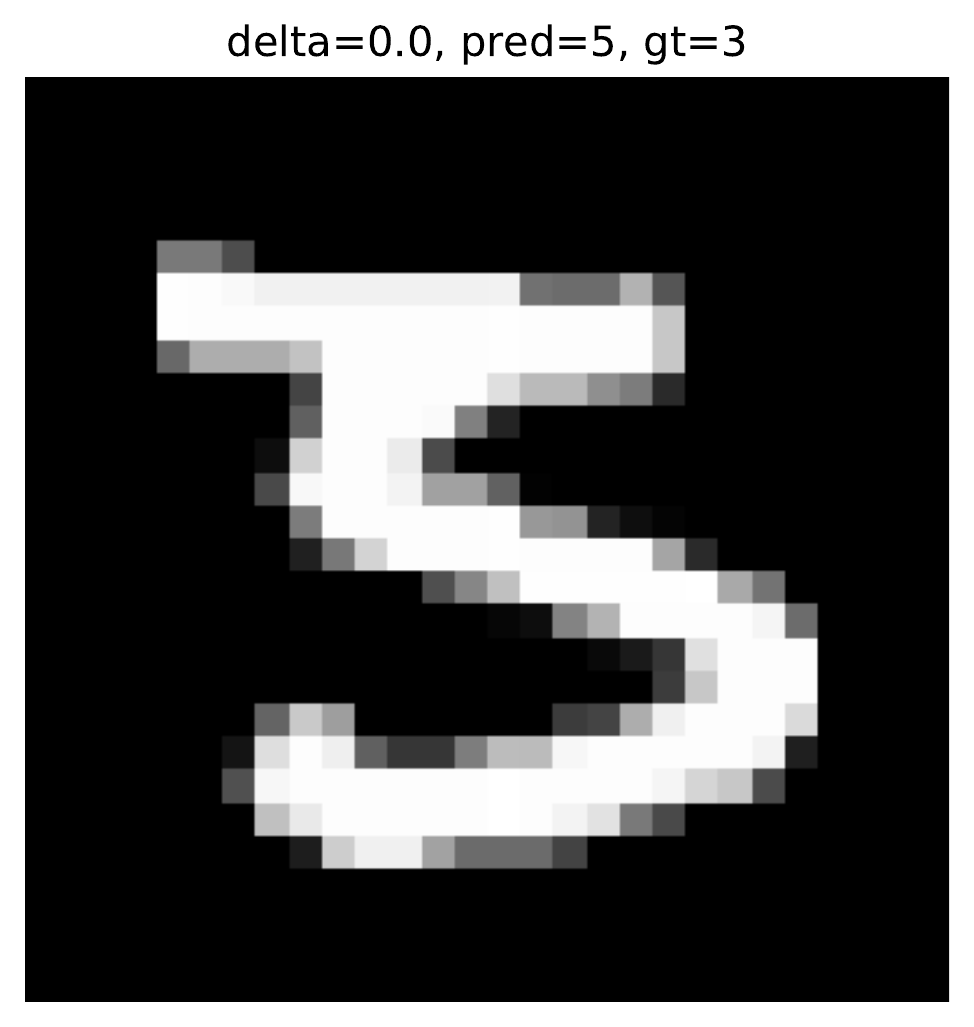}
         \caption{A testing image from MNIST with ground truth 3, classified as 5}
         \label{fig:misclsfy_3_to_5}
     \end{subfigure}
     \hfill
     \begin{subfigure}[t]{0.3\textwidth}
         \centering
         \includegraphics[width=\textwidth]{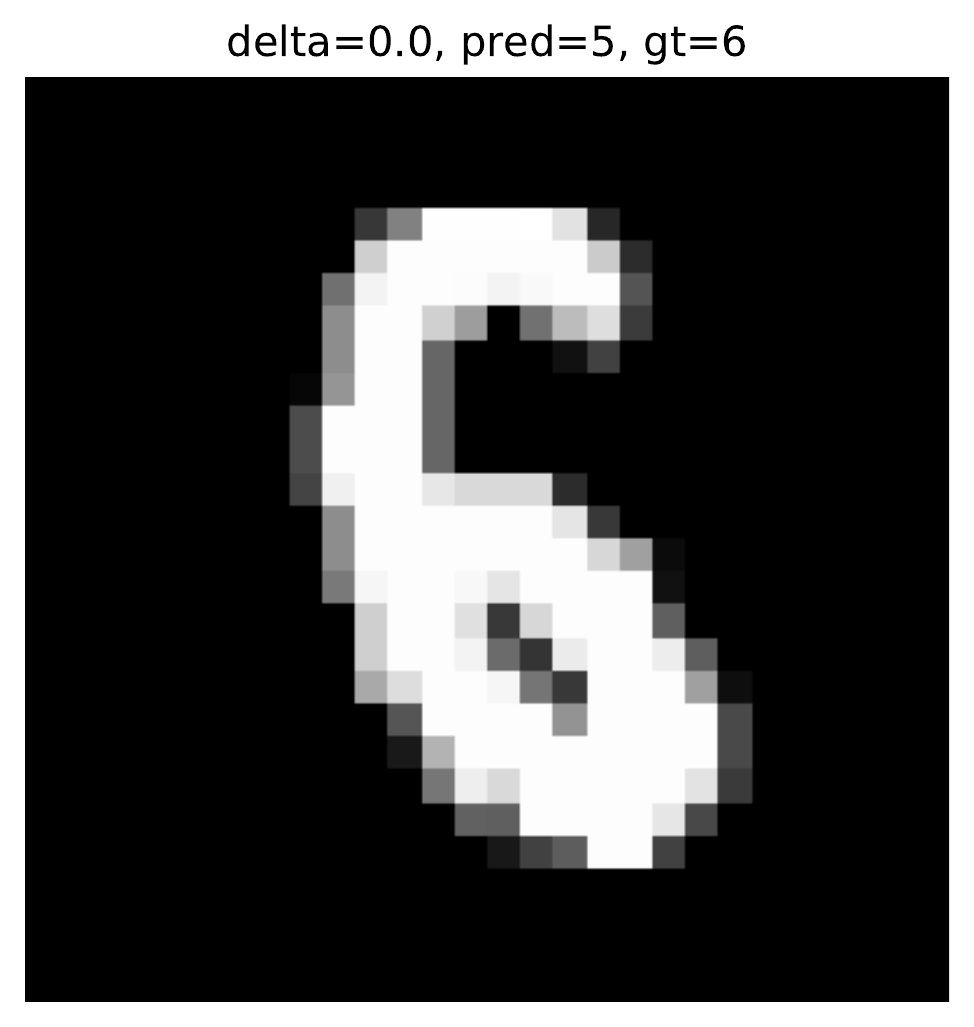}
         \caption{A testing image from MNIST with ground truth 6, classified as 5}
         \label{fig:misclsfy_6_to_5}
     \end{subfigure}
     \hfill
     \begin{subfigure}[t]{0.3\textwidth}
         \centering
         \includegraphics[width=\textwidth]{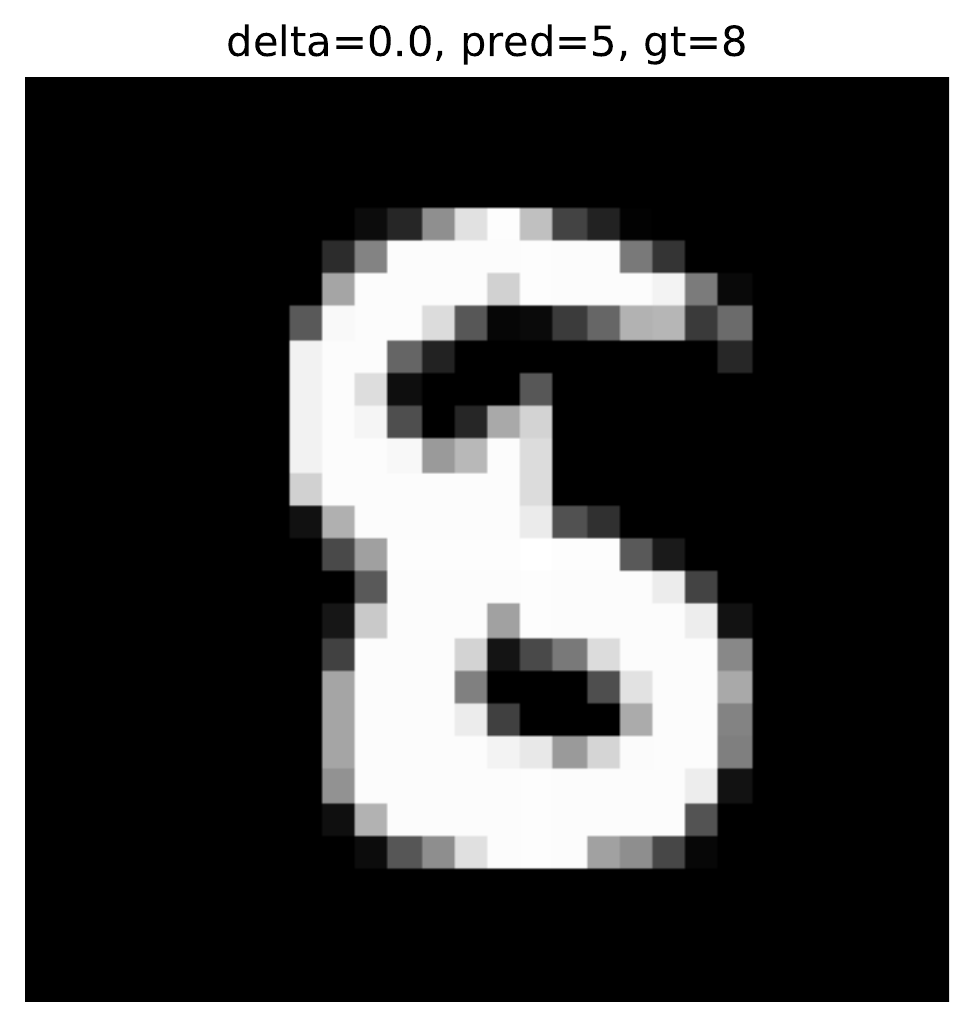}
         \caption{A testing image from MNIST with ground truth 8, classified as 5}
         \label{fig:misclsfy_8_to_5}
     \end{subfigure}
        \caption{Some interesting test images from MNIST that are misclassified as 5 and also follow the $\NAP$ of class 5.}
        \label{fig:Misclassify_to_5}
\end{figure}
\begin{figure}[th]
     \centering
     \begin{subfigure}[t]{0.3\textwidth}
         \centering
         \includegraphics[width=\textwidth]{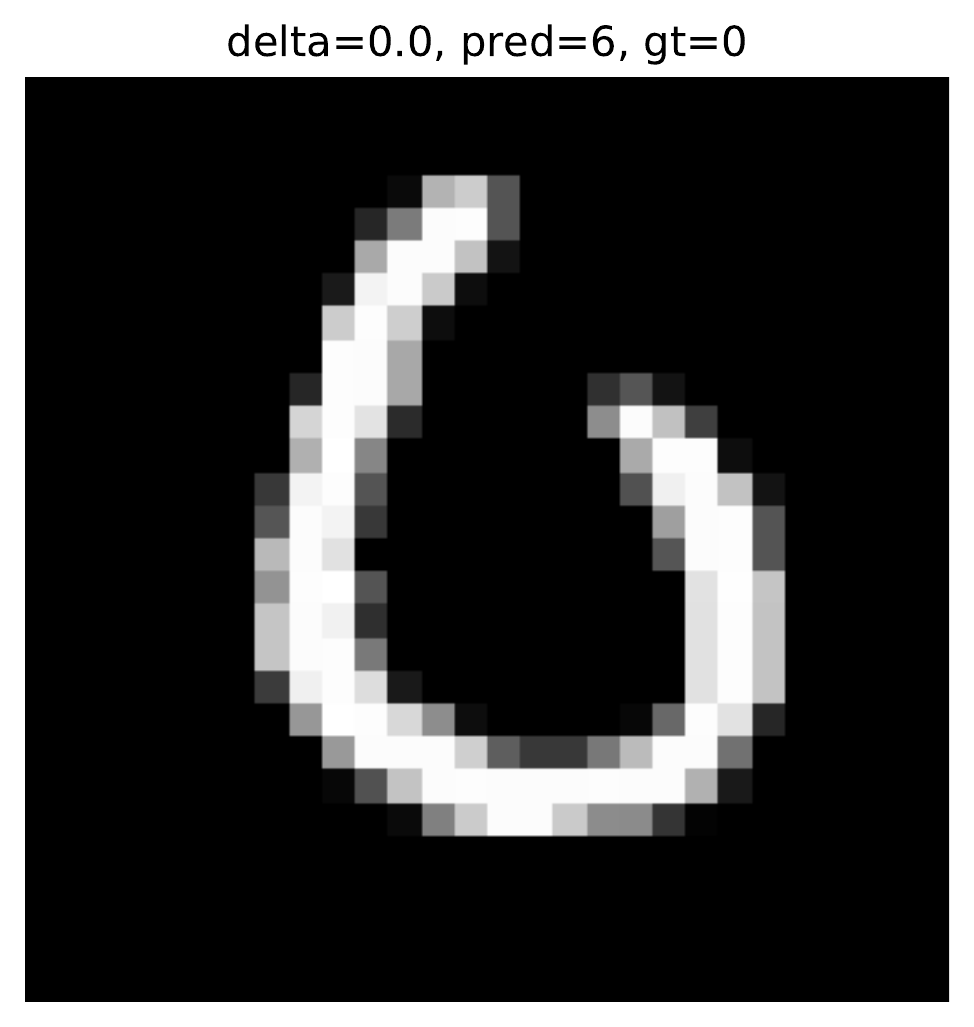}
         \caption{A testing image from MNIST with ground truth 0, classified as 6}
         \label{fig:misclsfy_0_to_6}
     \end{subfigure}
     \hfill
     \begin{subfigure}[t]{0.3\textwidth}
         \centering
         \includegraphics[width=\textwidth]{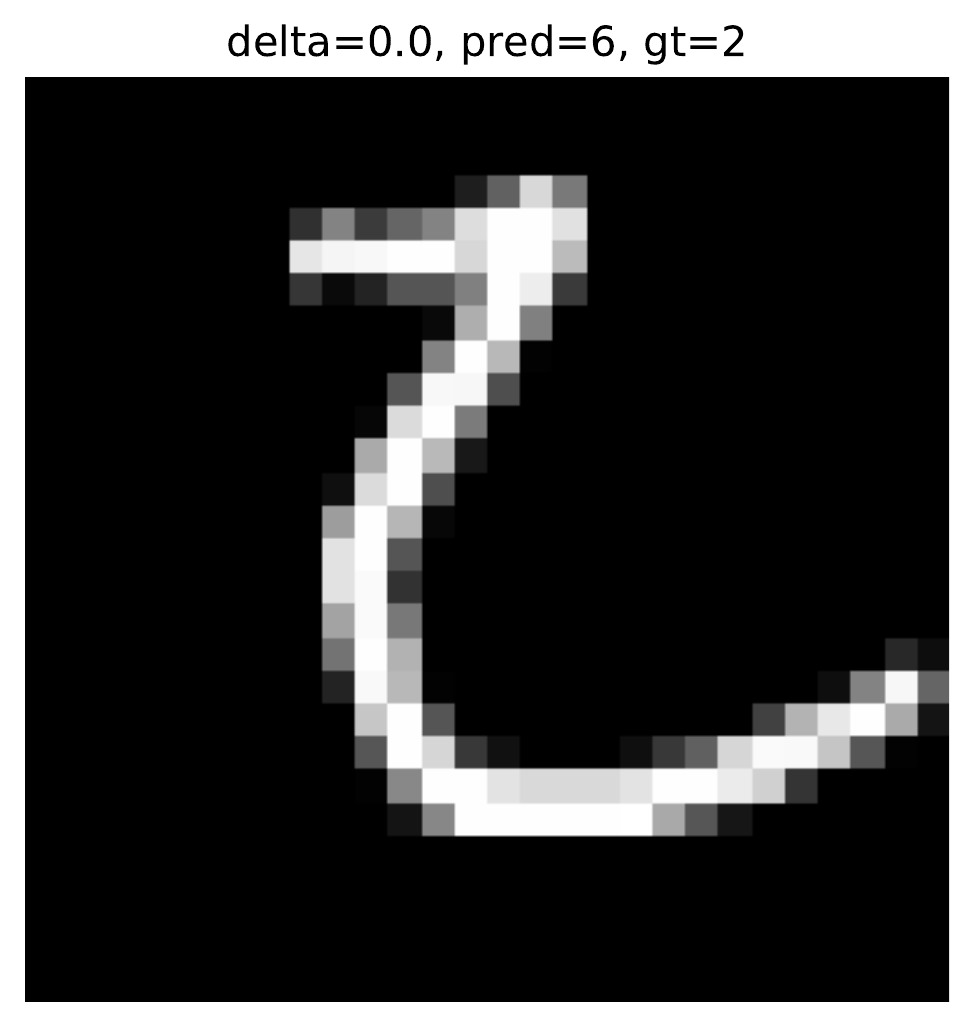}
         \caption{A testing image from MNIST with ground truth 3, classified as 6}
         \label{fig:misclsfy_2_to_6}
     \end{subfigure}
     \hfill
     \begin{subfigure}[t]{0.3\textwidth}
         \centering
         \includegraphics[width=\textwidth]{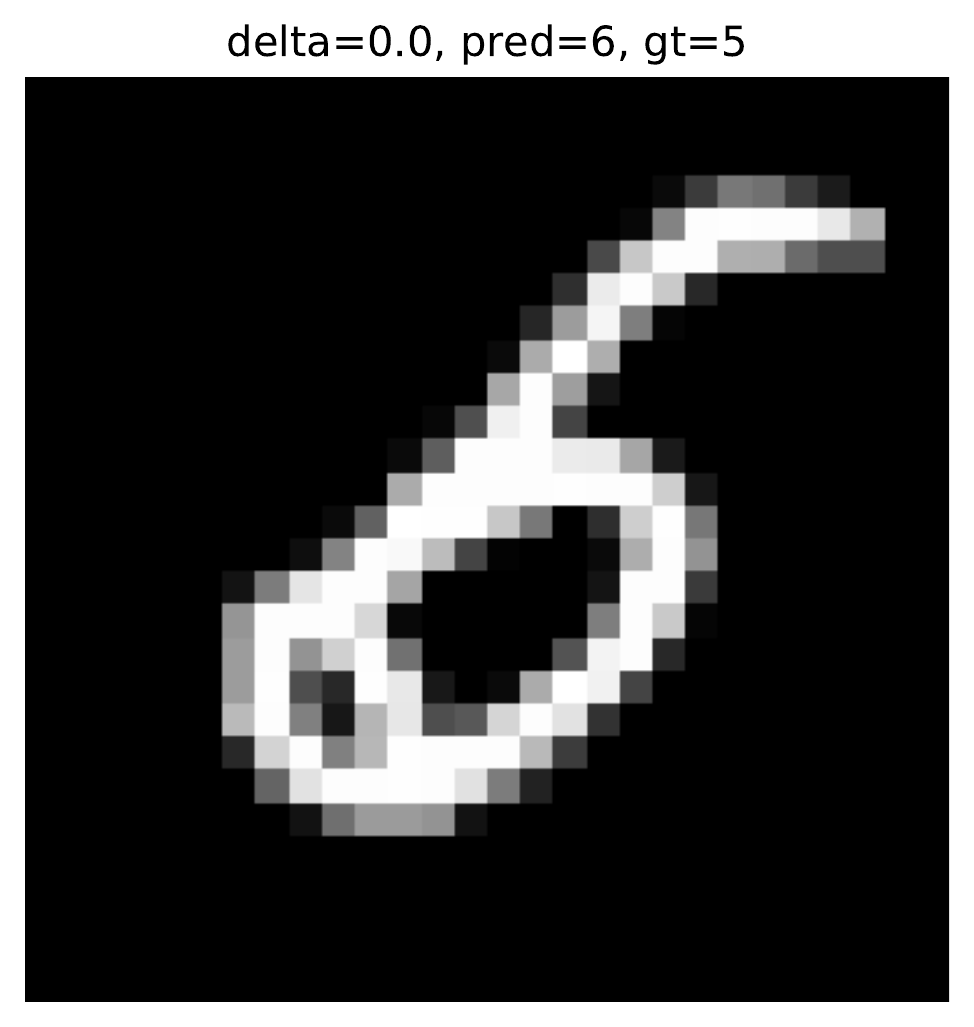}
         \caption{A testing image from MNIST with ground truth 5, classified as 6}
         \label{fig:misclsfy_5_to_6}
     \end{subfigure}
        \caption{Some interesting test images from MNIST that are misclassified as 6 and also follow the $\NAP$ of class 6.}
        \label{fig:Misclassify_to_6}
\end{figure}
\begin{figure}[th]
     \centering
     \begin{subfigure}[t]{0.3\textwidth}
         \centering
         \includegraphics[width=\textwidth]{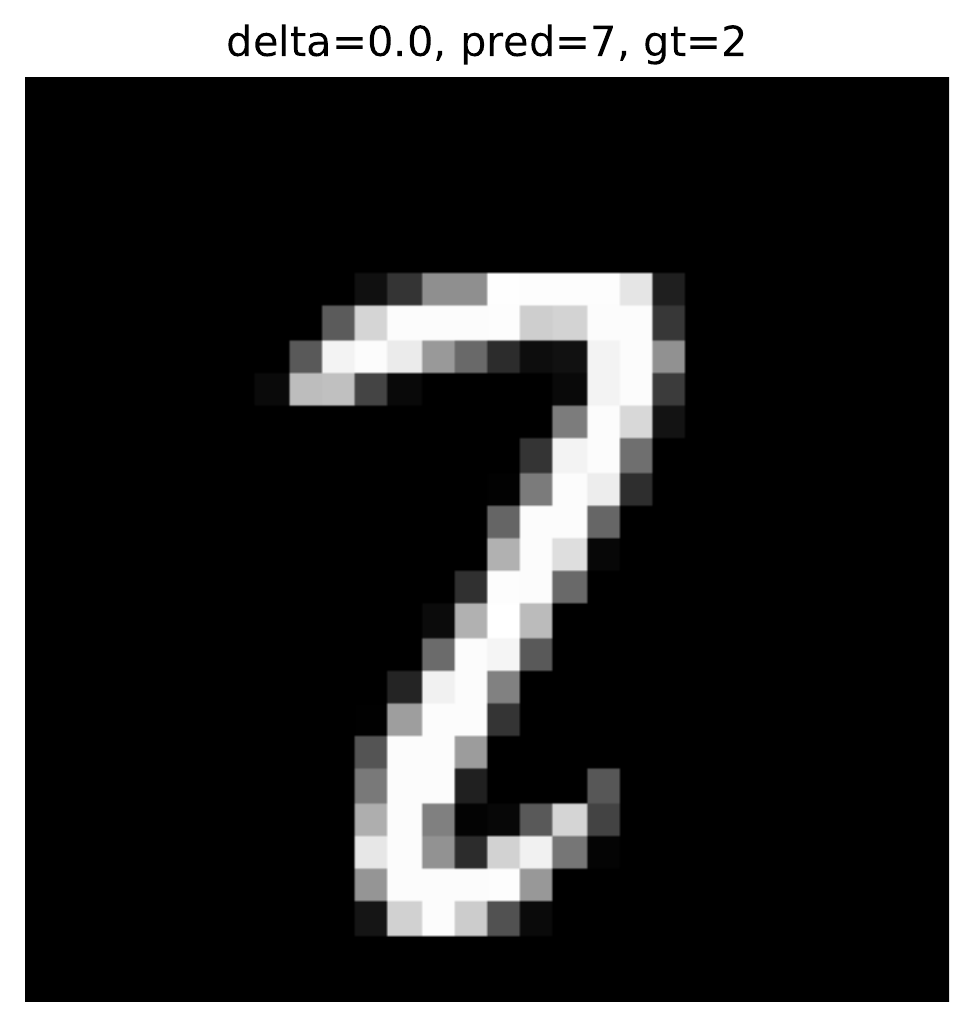}
         \caption{A testing image from MNIST with ground truth 2, classified as 7}
         \label{fig:misclsfy_2_to_7}
     \end{subfigure}
     \hfill
     \begin{subfigure}[t]{0.3\textwidth}
         \centering
         \includegraphics[width=\textwidth]{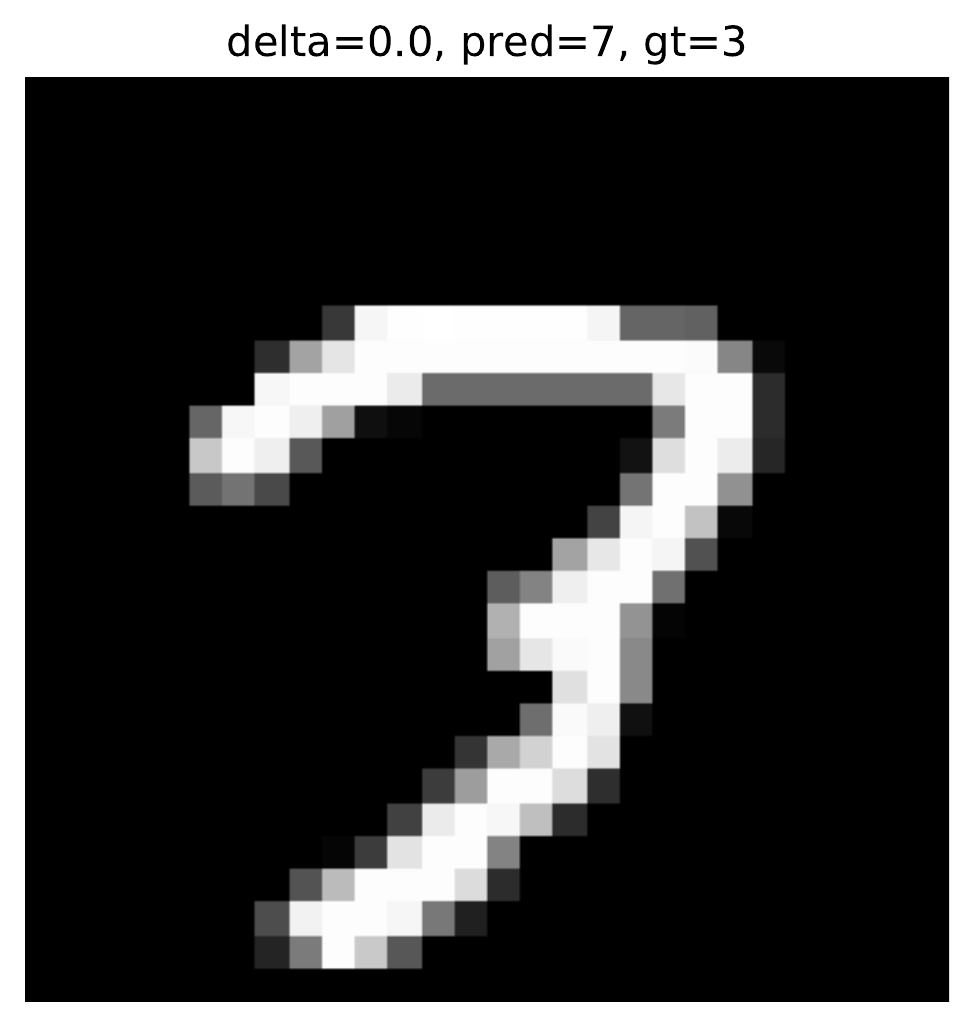}
         \caption{A testing image from MNIST with ground truth 3, classified as 7}
         \label{fig:misclsfy_3_to_7}
     \end{subfigure}
     \hfill
     \begin{subfigure}[t]{0.3\textwidth}
         \centering
         \includegraphics[width=\textwidth]{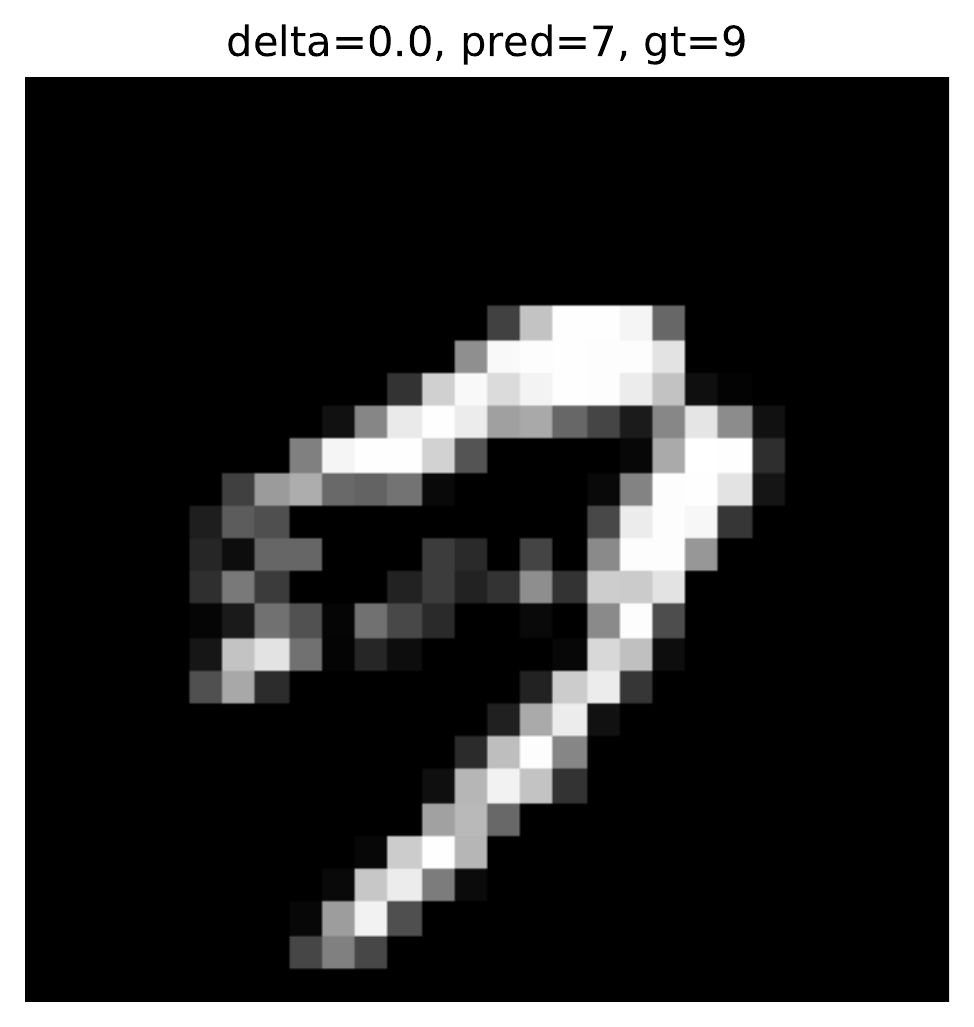}
         \caption{A testing image from MNIST with ground truth 9, classified as 7}
         \label{fig:misclsfy_9_to_7}
     \end{subfigure}
        \caption{Some interesting test images from MNIST that are misclassified as 7 and also follow the $\NAP$ of class 7.}
        \label{fig:Misclassify_to_7}
\end{figure}
\begin{figure}[th]
     \centering
     \begin{subfigure}[t]{0.4\textwidth}
         \centering
         \includegraphics[width=\textwidth]{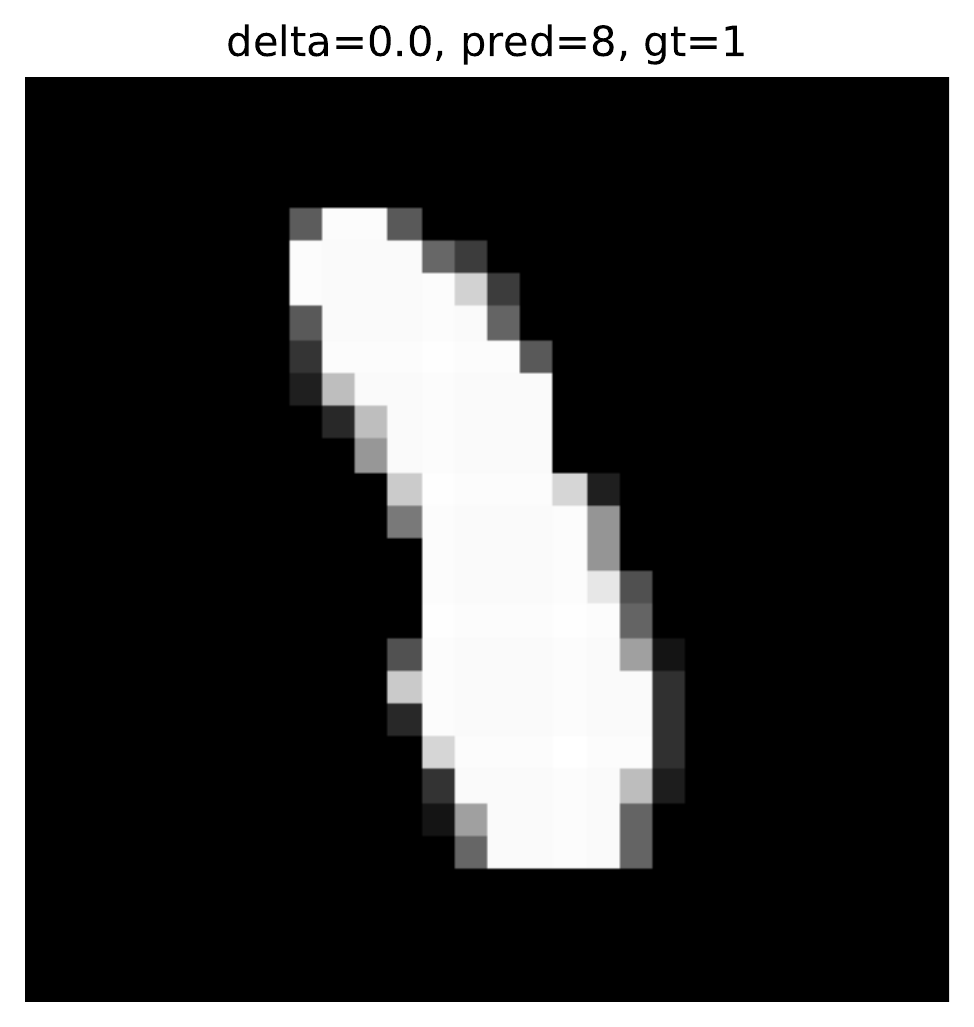}
         \caption{A testing image from MNIST with ground truth 1, classified as 8}
         \label{fig:misclsfy_1_to_8}
     \end{subfigure}
     \hfill
     \begin{subfigure}[t]{0.4\textwidth}
         \centering
         \includegraphics[width=\textwidth]{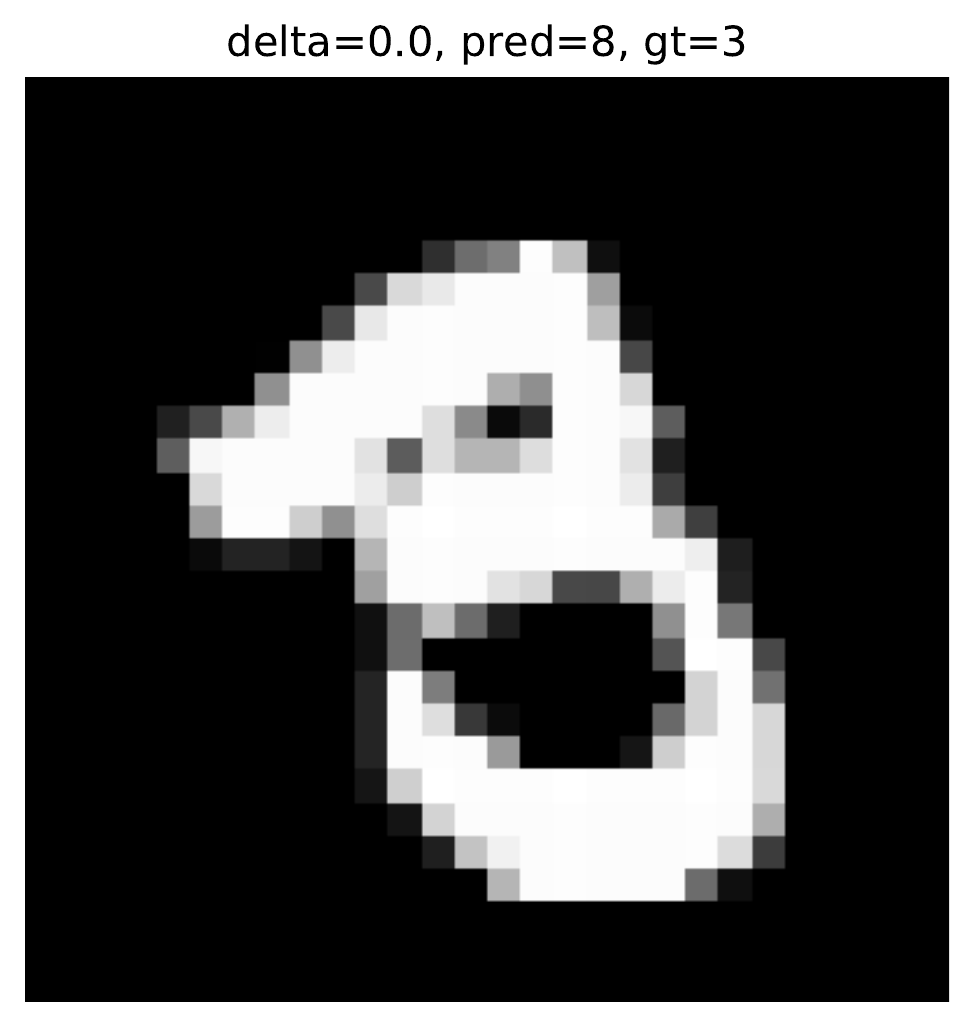}
         \caption{A testing image from MNIST with ground truth 3, classified as 8}
         \label{fig:misclsfy_3_to_8}
     \end{subfigure}
     \hfill
     % \begin{subfigure}[t]{0.3\textwidth}
     %     \centering
     %     \includegraphics[width=\textwidth]{figures/mnist_interesting_samples/eps0-0_pred7_gt9_idx980.pdf}
     %     \caption{A testing image from MNIST with ground truth 9, classified as 7}
     %     \label{fig:misclsfy_9_to_7}
     % \end{subfigure}
        \caption{Some interesting test images from MNIST that are misclassified as 8 and also follow the $\NAP$ of class 8.}
        \label{fig:Misclassify_to_8}
\end{figure}
\begin{figure}[th]
     \centering
     \begin{subfigure}[t]{0.4\textwidth}
         \centering
         \includegraphics[width=\textwidth]{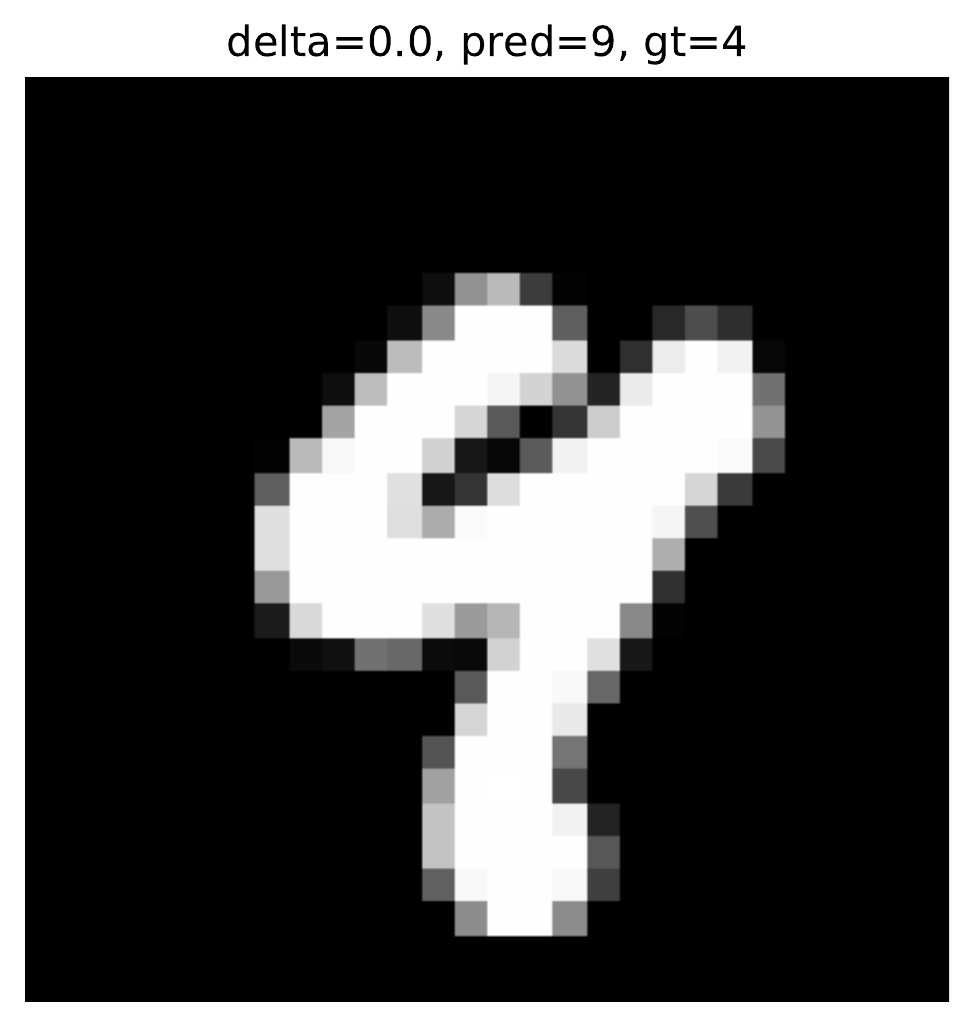}
         \caption{A testing image from MNIST with ground truth 4, classified as 9}
         \label{fig:misclsfy_4_to_9}
     \end{subfigure}
     \hfill
     \begin{subfigure}[t]{0.4\textwidth}
         \centering
         \includegraphics[width=\textwidth]{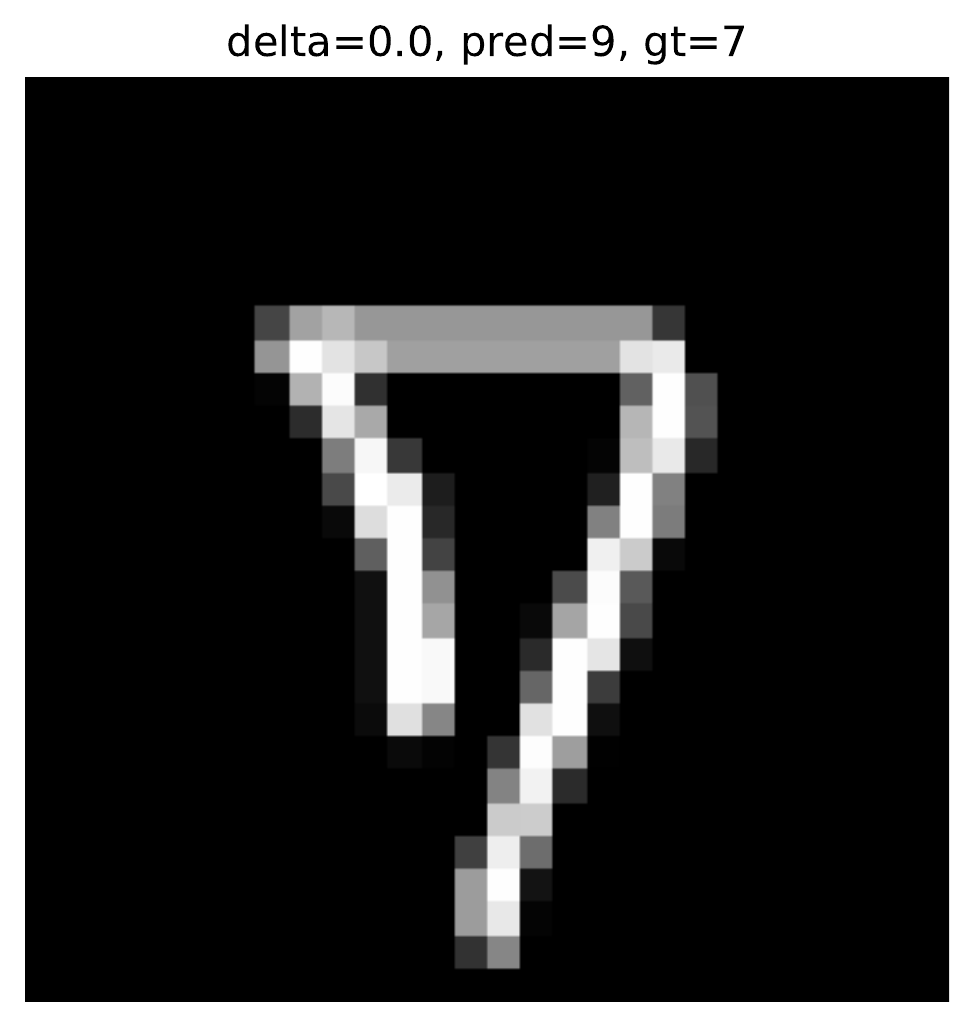}
         \caption{A testing image from MNIST with ground truth 7, classified as 9}
         \label{fig:misclsfy_7_to_9}
     \end{subfigure}
     % \hfill
     % \begin{subfigure}[t]{0.3\textwidth}
     %     \centering
     %     \includegraphics[width=\textwidth]{figures/mnist_interesting_samples/eps0-0_pred7_gt9_idx980.pdf}
     %     \caption{A testing image from MNIST with ground truth 9, classified as 7}
     %     \label{fig:misclsfy_9_to_7}
     % \end{subfigure}
        \caption{Some interesting test images from MNIST that are misclassified as 9 and also follow the $\NAP$ of class 9.}
        \label{fig:Misclassify_to_9}
\end{figure}

\end{document}